\documentclass[fleqn,10pt]{wlscirep}
\usepackage[utf8]{inputenc}
\usepackage{hyperref}
\usepackage{multirow}
\usepackage{longtable}
\usepackage{pdflscape}
\usepackage{graphicx}
\usepackage{caption}
\usepackage{amsmath}
\usepackage{amssymb}
\usepackage{booktabs}
\usepackage{xcolor}
\usepackage{colortbl}
\usepackage{bbding}
\usepackage{pifont}
\definecolor{tablehead}{RGB}{222,220,204}
\definecolor{tablesection}{RGB}{238,236,224}
\definecolor{tablesubtotal}{RGB}{247,246,240}

\renewcommand{\thesection}{\Roman{section}}
\renewcommand{\thesubsection}{\thesection.\arabic{subsection}}

\title{RadOnc-Agent: An Integrated Framework for AI Workflows Across the
Radiotherapy Care Pathway}

\author[1]{Caiwen Jiang}
\author[1]{Shuoyang Wei}
\author[1]{Songlin Zhao}
\author[1]{Junyu Li}
\author[1]{Jingyuan Chen}
\author[1]{Yuzhen Ding}
\author[2]{Daniel J. Ma}
\author[1]{Samir H. Patel}
\author[1]{Steven E. Schild}
\author[1]{Sujay A. Vora}
\author[1,*]{Wei Liu}

\affil[1]{Department of Radiation Oncology, Mayo Clinic Arizona, Phoenix, AZ, USA.}
\affil[2]{Department of Radiation Oncology, Mayo Clinic, Rochester, MN, USA.}
\affil[*]{liu.wei@mayo.edu}

\begin{abstract}
Artificial intelligence has advanced radiotherapy tasks, but these
capabilities remain fragmented across the care pathway. Here we present
RadOnc-Agent, an integrated framework connecting 17 phase-specific functions
and 9 cross-cutting utilities across four phases through a
conversational interface. A language-model controller maps clinical intent
to structured calls, maintains longitudinal context and validates dispatch.
We evaluated request-expression diversity, function coverage, cross-stage
composition, information transfer and failure recovery using 2,600
single-function requests (7,800 executions), 200 cross-stage
scenarios (600 executions), and 120 workflows from 60
de-identified patient records (360 executions). RadOnc-Agent selected
the intended function in 98.79\% of single-function executions and completed
96.50\% of scripted and 96.67\% of real-patient workflows. Removing
longitudinal state reduced cross-stage completion from 96.50\% to 84.00\%;
bypassing schema and identity validation increased mismatched dispatch from
0\% to 95.28\% in replay/test execution. Single-function performance did not
by itself predict workflow-level capability. Information continuity and
cross-stage dependencies emerged as central constraints. These results
establish the technical feasibility of longitudinal integration across
radiotherapy workflows and provide a foundation for prospective clinical
evaluation.
\end{abstract}


\begin{document}
\maketitle

%% =======================================================
\section{Introduction}
%% =======================================================

Radiotherapy is a longitudinal care pathway spanning diagnosis, patient
evaluation and treatment decision-making, simulation and planning, image-guided
delivery and adaptation, and post-treatment surveillance
~\cite{huynh2020artificial,kawamura2024revolutionizing}.
Information generated at one stage constrains the next. Intent and prescription
determine delineation requirements, anatomy and beam geometry constrain dose,
changes during treatment may prompt replanning, and delivered dose informs
toxicity assessment. These dependencies link clinical decisions to imaging,
structures, plans, delivered treatment and subsequent outcomes, across
different software environments and professional roles. Radiotherapy
intelligence must therefore operate across connected decisions, images,
quantitative models and longitudinal records.

Artificial intelligence has nevertheless advanced mainly through task-specific
models. Systems can delineate organs at risk
~\cite{totalsegmentator,zhang2024promptableseg}, predict
three-dimensional dose~\cite{zhang2023beammask}, register longitudinal images
~\cite{nenoff2023,ding2023dir} and synthesize computed tomography
~\cite{huijben2024synthrad,villegas2024syntheticct}.
Most remain point solutions with distinct inputs, outputs and software
environments, requiring users to select models, assemble inputs and transfer
outputs between systems. Their results must also be interpreted in light of
upstream intent and downstream consequences. Individual model performance
therefore does not by itself create a coherent radiotherapy workflow. The
unresolved problem is whether independently developed capabilities can be made
jointly accessible and reusable while preserving the clinical dependencies
that connect successive stages of care. This pathway-level question differs
from optimizing any one constituent function in clinical isolation.

Agentic artificial intelligence offers a potential control layer. Large
language models can translate natural-language intent into structured actions
and invoke external tools~\cite{yao2023react,schick2023toolformer}. Medical
agents and interactive benchmarks have retrieved longitudinal information and
evaluated sequential tasks in controlled clinical environments
~\cite{mira2026,schmidgall2026agentclinic};
radiotherapy agents have performed inverse planning or automated treatment
planning~\cite{nusrat2025dola,yang2026referencefree,
maniscalco2026compound,abrar2026m2m}, or coordinated functions within online
adaptive planning~\cite{arpa2026}. These systems establish coordination
within individual clinical episodes. They do not address whether a shared
framework can preserve treatment intent, patient identity and generated
artefacts as tasks progress from diagnosis and patient evaluation through
treatment decision-making, planning, adaptation and follow-up. Their
clinical-stage coverage, function types,
cross-stage composition, contextual continuity and patient-data evaluation are
compared in Table~\ref{tab:prior-agent-scope}.

A full-pathway framework requires a representation in which heterogeneous
clinical, imaging, dosimetric and retrieval capabilities are jointly
accessible, outputs can be reused downstream and patient context remains
continuous. This is an integration problem rather than a request for one model
to reproduce all specialist tasks. The language model can interpret requests
and select functions, while specialist models and services retain the imaging,
dosimetric and quantitative computations for which they were developed. Here
we present RadOnc-Agent, which organizes 17 phase-specific functions and 9
cross-cutting utilities into four radiotherapy phases
(Fig.~\ref{fig:overview}a). The core functions span treatment decisions,
planning, adaptation and outcomes, while utilities provide information access,
logistics and deterministic calculations. We ask whether this breadth remains
accessible and composable when information and prerequisites pass between
stages, rather than whether the controller itself can replace specialist
intelligence.
Flexibility is evaluated in terms of five measurable properties, namely request-expression
diversity, function coverage, cross-stage composition, information transfer and
failure recovery. We test these properties using single-function requests,
prespecified cross-stage scenarios and retrospective offline execution of two
real-patient workflows. Routing, state and validation ablations examine
mechanisms supporting execution, while function-specific evidence is kept
separate because heterogeneous task metrics cannot be pooled into end-to-end
clinical performance. This design assesses whether users can
reach each capability, whether capabilities can be connected while retaining
required information, and whether those connections execute on linked
patient-specific radiotherapy objects. By combining controlled synthetic
benchmarks with retrospective patient-specific execution, this design jointly
evaluates the breadth, composability and patient-specific execution of the
integrated framework.

%% =======================================================
\section{Results}
%% =======================================================

\subsection{An integrated function map spans the radiotherapy care pathway}
\label{sec:results-workflow-map}
RadOnc-Agent comprised 26 callable functions organized into a phase-specific
core layer and a cross-cutting utility layer (Fig.~\ref{fig:overview}a). The
17 core functions included two for Diagnosis, Patient Evaluation \&
Decision-Making, six for Simulation \& Planning, two for On-Treatment
Adaptation and seven for Follow-up \& Outcomes; nine utilities supported
retrieval, workflow logistics,
calculation and platform guidance across phases. Collectively, they spanned
treatment indication and modality selection, image processing, contouring,
dose and plan generation, adaptation and longitudinal outcome assessment.
Each capability had defined inputs and structured outputs, allowing individual
execution and downstream reuse within a shared patient context
(Fig.~\ref{fig:overview}b,c). Function-specific evidence is reported in
Supplementary Tables S1 and S2; workflow roles and benchmark exposure are
mapped in Extended Data Table~\ref{tab:function-workflow-coverage}. This map,
together with the per-function benchmark, defines function coverage at the
framework level.

\subsection{A unified interface supports diverse requests across heterogeneous specialist capabilities}
\label{sec:results-agent-architecture}
The shared interface linked the language-model controller, function registry
and specialist layer (Fig.~\ref{fig:overview}b). The controller interpreted a
request and patient context, selected a registered function and routed it to a
specialist model, retrieval or calculation service, or EHR-based utility.
Common callable contracts defined required inputs and outputs, and returned
structured results to the maintained workflow context. This separated
coordination from individual function implementations. Computationally
intensive imaging and dosimetric functions executed asynchronously, whereas
retrieval, calculation and EHR utilities returned synchronously through the
same interface; both modes retained provenance for downstream calls.
The interface thus allowed outputs from one function to become validated
inputs to another without requiring the controller to reproduce the
specialist computation.

The architecture benchmark comprised 2,600 unique requests (100 per function),
with 30 direct, 30 clinical-context, 20 paraphrased or reordered and 20 confusable
requests for each function. Three context-reset executions per request yielded
7,800 executions (Fig.~\ref{fig:architecture-evaluation}a).
RadOnc-Agent selected the intended function in 7,706 executions (98.79\%;
95\% CI, 98.46--99.10\%).
The generated call conformed to the schema of the invoked function in 7,743
executions (99.27\%; 95\% CI, 99.00--99.51\%), technical backend dispatch
succeeded in 7,719 (98.96\%; 95\% CI, 98.64--99.26\%), and 7,694 returned a
parseable structured output (98.64\%; 95\% CI, 98.28--98.97\%)
(Fig.~\ref{fig:architecture-evaluation}b). Across all four endpoints,
per-function success remained 98.00--99.67\%, indicating consistent
performance across the registry.

Function-selection accuracy was 99.79\% (2,335 of 2,340) for direct requests,
99.10\% (2,319 of 2,340) for clinical-context requests, 98.91\% (1,543 of
1,560) for paraphrased or reordered requests and 96.73\% (1,509 of 1,560) for
confusable requests
(Fig.~\ref{fig:architecture-evaluation}c). Confusable requests accounted for
51 of the 94 selection failures (54.3\%), making semantic overlap between
neighbouring functions the main source of routing error in this benchmark.
Across the 2,600
unique requests, the intended function was selected in all three sessions for
2,536 requests (97.54\%) and in at least two sessions for 2,579 (99.19\%);
three-of-three consistency ranged from 97.00\% to 98.33\% across function
groups (Fig.~\ref{fig:architecture-evaluation}d).

In 1,560 executions of 520 missing-input requests, missing information was
detected in 1,528
(97.95\%; 95\% CI, 96.99--98.85\%), and a safe clarification or withholding
response was issued in 1,516 (97.18\%; 95\% CI, 96.03--98.27\%); 44 executions
(2.82\%; 95\% CI, 1.79--3.97\%) inappropriately attempted to proceed
(Fig.~\ref{fig:architecture-evaluation}e). Group-level detection and safe
handling, defined as clarification or withholding without backend dispatch,
ranged from 97.78--98.33\% and 97.04--97.50\%, respectively.
Across the 7,719 normal executions that successfully reached backend dispatch,
median controller orchestration overhead ranged from 1.31 to 1.56~s across
function groups; the overall median was 1.42~s (interquartile range,
1.08--1.96~s), excluding backend inference time
(Fig.~\ref{fig:architecture-evaluation}f). These endpoints quantify
function selection, argument construction and execution reliability across the
integrated framework; all 95\% confidence intervals are reported in
Supplementary Table S3.

\subsection{Routing, state and validation support reliable integrated execution}
\label{sec:results-architecture-ablation}
We compared the complete framework with prespecified routing, state and
validation ablations using the same registry, inputs and scoring rules
(Extended Data Table~\ref{tab:architecture-ablation};
Fig.~\ref{fig:architecture-ablation}).
In the single-function benchmark, the full controller selected the intended
function in 7,706 of 7,800 executions (98.79\%), compared with 7,448
(95.49\%) for the embedding-based router (difference, +3.30 percentage
points). The difference increased from +1.28 percentage points for direct
requests to +7.31 percentage points for confusable requests
(Fig.~\ref{fig:architecture-ablation}a).

In the cross-stage benchmark, information was transferred correctly in 2,507
of 2,550 eligible
hand-offs (98.31\%) with the stateful controller and 2,269 (88.98\%) with the
stateless controller; corresponding context-retention rates were 4,782 of
4,800 turns (99.63\%) and 4,382 (91.29\%). The stateful controller generated
34 unnecessary calls among 3,150 required calls (1.08\%), compared with 128
(4.06\%) for the stateless controller (Fig.~\ref{fig:architecture-ablation}b).
These execution-level differences translated into completion of 579 of 600
workflows (96.50\%) and adherence to the reference trajectory in 570 (95.00\%)
with the stateful controller, compared with 504 (84.00\%) and 477 (79.50\%),
respectively, for the stateless controller
(Fig.~\ref{fig:architecture-ablation}c).

In replay or test execution, the enabled validation gates blocked all 44
controller attempts to dispatch a call with a required input missing (0 of
1,560 unsafe dispatches), whereas bypassing the gates permitted 44 unsafe
dispatches (2.82\%). In the 360 identifier-integrity executions, enabled gates
safely rejected 354 mismatches (98.33\%) after detecting 356 (98.89\%); the
bypassed configuration detected 19 (5.28\%) and safely rejected 17 (4.72\%).
No mismatched dispatch occurred with
enabled gates, whereas 343 of 360 (95.28\%) were dispatched when the gates
were bypassed
(Fig.~\ref{fig:architecture-ablation}d,e). No bypassed call was routed to an
actual registered service. Together, these comparisons identify complementary
contributions from language-based routing, persistent state and deterministic
validation; corresponding 95\% confidence intervals are reported in
Supplementary Table S4.

\subsection{RadOnc-Agent composes functions and transfers information across radiotherapy stages}
\label{sec:results-cross-stage}
The cross-stage benchmark comprised 200 prespecified scenarios (50 in each of
four workflow families), each executed three times in context-reset sessions
(600 executions). Ninety executions received a missing-input or
transient-backend perturbation. Clinical inputs were released by turn; the
expected functions and hidden reference trajectory were used only for scoring
(Fig.~\ref{fig:e2e}a). The agent received clinical objectives but not the
function names or order encoded in that trajectory. RadOnc-Agent completed all
required calls in 579 of 600
executions (96.50\%) and adhered to the reference trajectory in 570 (95.00\%).
Information was transferred correctly in 2,507 of 2,550 hand-offs (98.31\%),
and patient and treatment context was retained in 4,782 of 4,800 turns
(99.63\%) (Fig.~\ref{fig:e2e}b). No cross-case contamination was observed; the
18 context-retention errors occurred within the active case. These endpoints
therefore test whether outputs and patient-specific state remained available
to downstream functions, beyond selection of isolated tools.

Completion was 148 of 150 for decision-to-planning workflows (98.67\%), 145 of
150 for planning-to-adaptation (96.67\%), 147 of 150 for
treatment-to-follow-up (98.00\%) and 139 of 150 for full-pathway workflows
(92.67\%) (Fig.~\ref{fig:e2e}b). All three repetitions completed for 48, 46,
47 and 43 of 50 scenarios in these respective families; the corresponding
counts for trajectory adherence were 47, 44, 46 and 43
(Fig.~\ref{fig:e2e}c). The lower full-pathway completion identified longer
call chains as the more demanding setting. Completion was 343 of 360
(95.28\%) under registered-service execution and 236 of 240 (98.33\%) under
replay/test execution; trajectory adherence was 93.89\% and 96.67\%,
respectively (Fig.~\ref{fig:e2e}d). Performance was stratified descriptively by
service mode to determine whether coordination was retained when the framework
interacted with registered specialist services as well as controlled replay
responses.

Of 21 incomplete workflows, nine reflected failed upstream-output propagation,
six a prerequisite or ordering violation, four persistent backend failure and
two unresolved missing inputs (Fig.~\ref{fig:e2e}e). Among 90 perturbed
executions, 84 recovered and completed the workflow (93.33\%). The agent made
34 unnecessary calls relative to 3,150 required calls (1.08\%) and no calls
containing information from another case across 4,800 evaluated turns
(Fig.~\ref{fig:e2e}f). The concentration of failures in propagation and
ordering identifies information continuity and cross-stage dependency
management as the principal constraints observed. Required trajectories
operationalized composition, eligible hand-offs and scored turns measured
information transfer, and perturbed executions tested recovery.

\subsection{Integrated workflows execute on retrospective real-patient data}
\label{sec:results-clinical-validation}
We conducted a retrospective, offline workflow-execution evaluation using
60 de-identified real-patient radiotherapy records that were held out from
framework development and model training. The cohort contained 12 patients
from each of five anatomical groups. Each contributed decision-to-planning and
planning-to-adaptation workflows (120 instances), executed three times in
context-reset sessions (360 clean executions). A separate 48-execution stress
test introduced missing inputs or transient backend failures. Source records
linked diagnosis and course metadata to planning images, structures, plan and
dose objects, and on-treatment verification imaging; the controller, registry
and specialist functions were held fixed during evaluation. The two workflow
families were selected because these linked objects permit patient-specific
testing of both an upstream planning transition and a downstream adaptive
transition. These workflows therefore sampled distinct data dependencies
rather than repeated variants of the same transition within the pathway alone.

Decision-to-planning supplied recorded intent, prescription and planning CT
and required a planning directive followed by delineation.
Planning-to-adaptation supplied the reference image, structures and dose with
verification imaging and required registration followed by replan assessment.
The recorded subsequent-plan-change endpoint provided an auditable
longitudinal marker of treatment modification after the selected verification
CT. The endpoints were selected to quantify patient-specific workflow
composition, identifier continuity, structured information transfer and
execution reliability using auditable source objects.

Across 360 clean workflow executions, RadOnc-Agent completed all required
functions in 348 (96.67\%) and adhered to the prespecified trajectory in 344
(95.56\%) (Fig.~\ref{fig:clinical-evaluation}b). Among 720 required function
calls, 713 (99.03\%) had schema-valid arguments, 711 (98.75\%) were
successfully dispatched and 708 (98.33\%) returned parseable structured
outputs. Patient, course and plan identifiers were propagated correctly in 351
of 354 eligible hand-offs (99.15\%).
Completion was 176 of 180 (97.78\%) for decision-to-planning workflows and
172 of 180 (95.56\%) for planning-to-adaptation workflows; the corresponding
trajectory-adherence rates were 174 of 180 (96.67\%) and 170 of 180
(94.44\%), respectively (Fig.~\ref{fig:clinical-evaluation}c). Context was
retained throughout 355 of 360 executions (98.61\%), with no cross-patient
contamination observed. The agent issued 10 unnecessary calls relative to 720
required calls, corresponding to an excess-call ratio of 1.39\%
(Fig.~\ref{fig:clinical-evaluation}d). Median controller overhead was 1.47~s
(interquartile range, 1.10--2.03~s), excluding specialist-model inference.

In the stress test, 45 of 48 executions recovered (93.75\%). Recovery after a
missing input was 12 of 12 for decision-to-planning and 11 of 12 for
planning-to-adaptation; recovery after a transient backend failure was 11 of
12 for each workflow family (Fig.~\ref{fig:clinical-evaluation}e). Across clean
runs, 112 of 120 workflows (93.33\%) completed in all three, five in two, two
in one and one in none. Three-of-three completion was achieved for 57 of 60
decision-to-planning and 55 of 60 planning-to-adaptation workflows; 59 and 58,
respectively, completed in at least two runs (Fig.~\ref{fig:clinical-evaluation}f).
Among the 60 planning-to-adaptation workflows, 24 source records contained a
subsequently recorded plan change. The replan-assessment function emitted a
subsequent-plan-change flag in 22 of these 24 workflows and correctly did not
emit the flag in 33 of 36 workflows without a subsequently recorded plan
change, yielding overall agreement of 55 of 60 (91.67\%), precision of 22 of
25 (88.00\%) and an F1 score of 89.80\%
(Fig.~\ref{fig:clinical-evaluation}g). This comparison quantifies alignment
with the longitudinal treatment-record endpoint.

\subsection{Task-specific evidence spans the integrated specialist capabilities}
\label{sec:results-specialist-evidence}
The specialist layer was evaluated function by function using task-appropriate
reference standards and native endpoints. Representative studies span
guideline-grounded decision support, image synthesis, registration,
auto-contouring, dose prediction, plan optimization, adaptive imaging and
longitudinal outcome assessment. Selected native endpoints are shown in
Extended Data Fig.~\ref{fig:specialist-evidence}.
The representative results show performance at the native scale of each task,
including treatment-decision endpoints across two configurations, anatomical
overlap and deformation quality for registration, site-stratified contouring,
in-domain and unseen-configuration dose prediction, dosimetric verification,
and physician-adjudicated outcome assessment. Published, internal and
system-level evidence is distinguished according to the implementation and
clinical role of each capability. Together, these evaluations establish the
breadth of specialist functions available for workflow composition; complete
function descriptions, cohorts, reference standards, native metrics and 95\%
confidence intervals are reported in Supplementary Tables S1, S2 and S5.

%% =======================================================
\section{Discussion}
%% =======================================================

RadOnc-Agent addresses the fragmentation of radiotherapy artificial
intelligence rather than introducing a new language model or another isolated
prediction algorithm. Its contribution is the integration of 17 phase-specific
functions and 9 cross-cutting utilities within a common representation of the
radiotherapy pathway, with the language model coordinating rather than
replacing specialist services. Flexibility was operationalized as request-
expression diversity, coverage, cross-stage composition, information transfer
and recovery. Function selection ranged from 96.73\% to 99.79\% across request
types; all 26 capabilities were exercised; scripted and real-patient workflow
completion was 96.50\% and 96.67\%; and most perturbed executions recovered.
Information transfer reached 98.31\% in hand-offs and identifier propagation
99.15\% in real-patient hand-offs; recovery was 93.33\% and 93.75\%,
respectively.
Together, these results demonstrate the reuse of heterogeneous specialist
capabilities within longitudinal radiotherapy workflows and identify
information continuity as a central determinant of successful composition.

The architecture ablations provide supporting evidence about the conditions
needed for reliable integration. Language-based routing had its largest
advantage for confusable requests, persistent state preserved information
between stages and deterministic validation blocked incomplete or
identity-inconsistent calls. Removing state reduced completion from 96.50\%
to 84.00\%, whereas bypassing validation permitted mismatched dispatch in
95.28\% of identifier-integrity executions. These comparisons identify
language-based routing, persistent state and deterministic validation as
complementary mechanisms supporting integrated execution. Function-level
evidence separately characterizes the constituent capabilities using
task-appropriate clinical, imaging, dosimetric or technical endpoints.

Longer workflows remained more difficult despite the breadth of integration.
Although intended-function selection reached 98.79\% in isolated requests,
completion fell to 92.67\% for synthetic full-pathway workflows, and failures
were concentrated in propagation of upstream outputs and prerequisite or
ordering violations. Removing longitudinal state further reduced overall
cross-stage completion from 96.50\% to 84.00\%. Taken together, these findings
show that workflow-level capability is not a simple extension of
single-function performance. Under the evaluated conditions, maintaining
information continuity and satisfying cross-stage dependencies emerged as
central constraints as the clinical trajectory lengthened. This divergence
makes information continuity, rather than isolated model accuracy, the central
systems-level finding of this study.

The real-patient evaluation focused on decision-to-planning and
planning-to-adaptation workflows, for which linked patient, imaging, structure,
plan and dose objects could be assembled reliably. Evaluating the complete
pathway requires longitudinal reconstruction of treatment decisions,
successive plan versions, fraction-level delivery records, adaptive events and
long-term follow-up outcomes. These data are distributed across the electronic
health record, oncology information system, imaging archive and
treatment-planning system, often with differing identifiers, timestamps and
documentation formats. Reliable linkage also requires manual review to
distinguish planned treatment transitions from adaptive changes and to
adjudicate longitudinal outcomes. Building a sufficiently complete and
well-curated full-pathway cohort is therefore time- and resource-intensive. We
are currently assembling such a dataset for future evaluation across all four
phases.

This work also differs in scope from existing medical and radiotherapy agents
(Table~\ref{tab:prior-agent-scope}). General tool-using language-model
approaches established that reasoning models can invoke external actions and
computational resources~\cite{yao2023react,schick2023toolformer}, but they are
not clinical workflow systems and are therefore not included in the table.
MIRA spans diagnostic and therapeutic actions within a general emergency-care
episode, whereas the radiotherapy systems in the comparison focus on inverse
planning or a daily adaptive-planning episode. RadOnc-Agent extends integration
across decision-making, simulation and planning, on-treatment adaptation, and
follow-up, with patient-, course- and plan-level context passed between stages.
Its advantage is therefore the breadth and continuity of the radiotherapy
workflow represented, rather than superior language-model performance.

The results also clarify how orchestration should operate within clinical
decision-making. RadOnc-Agent supports use under clinical supervision, with
its intermediate and final outputs subject to professional review. In the
systematic missing-input benchmark, the agent issued a safe clarification or
withholding response in 1,516 of 1,560 executions (97.18\%) and attempted to
proceed in 44 (2.82\%). Deterministic constraints therefore complement
language-model reasoning by enforcing input, identity and provenance
requirements and escalating insufficient evidence.

Several considerations define the scope of the present evidence. The
retrospective evaluation established workflow execution on linked
patient-specific objects. Prospective workflow studies can assess clinician
time, editing burden, overrides, acceptance and automation bias.
Function-specific evidence used task-appropriate datasets and endpoints
because the integrated capabilities span distinct clinical, imaging and
dosimetric tasks. External retrospective evaluation and prospective silent
deployment will assess transportability across institutions and local
workflows, while broader adversarial and multi-failure testing will extend
safety characterization. Consistent with guidance for radiotherapy and
clinical-AI evaluation
~\cite{hurkmans2024guideline,vasey2022decide,lekadir2025futureai}, these studies
will retain prespecified performance gates and professional review and progress
towards multidisciplinary human-in-the-loop assessment of clinical decisions,
efficiency and patient-centred outcomes.

%% =======================================================
\section{Conclusion}
%% =======================================================

RadOnc-Agent integrates 26 functions across the radiotherapy pathway.
Single-function, cross-stage and
retrospective real-patient evaluations demonstrated coverage, composition,
information transfer and recovery. Longitudinal
execution cannot be inferred from isolated-function performance alone.
Information continuity and cross-stage dependencies must be evaluated at the
workflow level. These findings establish a technical foundation for integrating
heterogeneous artificial-intelligence capabilities across the radiotherapy
care pathway. Ongoing full-pathway data collection and subsequent external and
prospective evaluation will extend this framework towards
clinician-supervised use.

%% =======================================================
\section{Methods}
%% =======================================================

\subsection{Ethics}
This single-site retrospective study at Mayo Clinic Arizona was approved by
the Mayo Clinic Institutional Review Board (IRB \#: 24-011106) and conducted
in accordance with the Declaration of Helsinki. The Institutional Review Board
waived the requirement for informed consent because of the retrospective study
design. Data access, storage and analysis were conducted within Mayo Clinic
Arizona in accordance with institutional data-use and security requirements.
The architecture and cross-stage system-engineering benchmarks used synthetic
scenario inputs only and were determined to be exempt from additional
human-subject research requirements.

\subsection{Study design, cohorts and data sources}
The study comprised four complementary evaluations, including a
single-function architecture benchmark of 2,600 unique requests, a scripted
cross-stage workflow benchmark of 200 unique scenarios, a retrospective
real-patient workflow-execution evaluation of 60 patients and 120 workflow
instances, and task-specific evaluations of the 26 callable functions.
For framework-level evaluation, technical workflow flexibility was defined by
five domains encompassing request-expression diversity, function coverage, cross-stage
composition, information transfer and recovery from missing inputs or transient
backend failures. Each domain was evaluated using endpoints from the
prespecified benchmarks rather than combined into a single flexibility score.
The architecture benchmark used a prespecified synthetic request set derived
from the registered function contracts, with 100 normal requests per function
(2,600 requests in total), supplemented by 20 missing-input requests per
function (520 requests in total). The cross-stage benchmark used 200
prespecified synthetic scenarios, with 50 scenarios in each of the
decision-to-planning, planning-to-adaptation, treatment-to-follow-up and
full-pathway workflow families. These two benchmarks used synthetic clinical
contexts and were designed to assess orchestration independently of
patient-level specialist-model performance.

The real-patient cohort was used to evaluate patient-specific execution of the
integrated framework on linked radiotherapy data, focusing on workflow
completion, information transfer, identifier continuity and recovery. Each
real-patient workflow was repeated
three times in independent context-reset sessions. Function-level evidence was
obtained separately from task-specific evaluations using retrospectively
collected clinical datasets and, where appropriate, public datasets. Because
the development and evaluation cohorts, reference standards and endpoints
differed across functions, concise function descriptions and headline results
are reported in Supplementary Table~S1; detailed
implementation, development and evaluation data are reported in Supplementary
Table~S2. Functions based on general-model
prompting, retrieval, deterministic calculation, database or API access, or
static responses did not require a task-specific model-training dataset. For
trained models, training, validation and evaluation datasets were separated at
the patient level unless otherwise specified; patient-specific and
external-validation designs are identified in Supplementary
Table~S2.

The real-patient cohort was retrospectively assembled from de-identified
radiotherapy planning and on-treatment imaging records at Mayo Clinic Arizona
from May 2016 through December 2024. Selection was based on three
considerations. First, each included treatment course had to contain the
clinical and technical information required by both evaluated workflows,
namely diagnosis and course metadata, treatment intent and prescription, planning CT,
structure set, treatment plan, dose distribution and at least one
on-treatment verification CT. Second, these objects had to be readable and
reliably linked to the same patient and treatment course through patient-,
course- and plan-level identifiers. Third, patients used for framework
development, prompt refinement or specialist-model development were excluded
before evaluation. Eligible records were then stratified into abdomen, breast,
head, head-and-neck and pelvic groups, and 12 patients meeting the same
criteria were included from each group. This balanced sampling was intended to
provide anatomical coverage for technical workflow evaluation rather than to
estimate the prevalence of disease sites in the source population. One
treatment course was selected per patient; each patient contributed two
workflow instances and was treated as a clustering unit in repeated-execution
analyses.

\subsection{System implementation and shared execution contract}
\label{sec:methods-system-implementation}
RadOnc-Agent comprises a language-model controller, a machine-readable
function registry, a dispatch layer and independently hosted backend
services. The controller's system prompt is compiled from the registry, which
specifies for each function its name and description, clinical scope, required
and optional arguments, output class, execution mode, backend route and
registered caveats. Given a user request and the available patient context,
the controller emits a structured function-call directive identifying the
selected function and its arguments. The dispatch layer validates the call
against the registered schema, invokes the corresponding backend and returns
the resulting payload to the conversational state.

The main application and computationally intensive specialist-model services
are implemented as independent Flask applications. Long-running inference,
including image synthesis, segmentation, registration and dose prediction,
uses an asynchronous job-queue pattern. A submission endpoint returns a job
identifier, a status endpoint reports progress and a download endpoint,
proxied by the main application, returns the completed result. Lightweight
knowledge retrieval, deterministic calculation and EHR-based utilities are
executed synchronously. Backend responses are normalized into a shared result
envelope before being added to the patient and workflow context, enabling a
subsequent function call to consume a preceding function's structured output.
For clinical information-access utilities, this integration layer enables
programmatic retrieval through the same callable contract, reducing the need
for manual navigation, copying and transfer across separate
oncology-information, electronic-health-record, imaging, messaging and
scheduling interfaces. The framework does not reinterpret a source-system
result when the registered utility is defined as direct retrieval.
The shared contract separates orchestration logic from model implementation,
so that an individual specialist service can be updated without changing the
controller or its function-calling protocol.

All 26 functions were implemented, registered and callable within
RadOnc-Agent. At the time of the study, text-only functions
were available to authorized clinical users. Some image-processing functions
requiring clinical image read or write access were restricted to retrospective
research and offline evaluation because the necessary clinical-system
permissions had not been enabled. These access levels were determined by
institutional data-governance and clinical-system permissions. In the
evaluations, \textit{registered-service} execution denotes
invocation of the actual registered computational or retrieval service; it
does not necessarily indicate that the function was enabled for routine
clinical use. Replay or test execution returned a controlled response
conforming to the same function contract.

All system-level evaluations used gpt-5.2 (OpenAI) as the controller model.
The controller configuration, 26-function registry and specialist backends
were held fixed throughout the system-level evaluations. Model calls used
fixed decoding settings (temperature 0.0 and top-$p$ 1.0) with automatic,
serial tool calling and a bounded retry policy for transient failures. The
system prompt and registry were unchanged during evaluation.
Patient and workflow context was maintained in a session-scoped state object
bound to one active patient and treatment course. Patient, course and plan
identifiers were checked before each backend dispatch, and calls inconsistent
with the active session were rejected. Structured outputs retained their
originating-function and patient-, course- and plan-level provenance. No
persistent conversational memory or cross-patient cache was used, and session
state was cleared after every independent execution. Clinical-backend
communication was encrypted and restricted to authenticated application
services, with audit logs retained for backend dispatches. The principal
execution risks, detection mechanisms and fail-safe responses are summarized
in Extended Data Table~\ref{tab:safety-controls}.

\subsection{Callable functions and implementation basis}
RadOnc-Agent comprised 17 phase-specific core functions and 9 cross-cutting
utilities. Core functions were assigned prospectively to one of four
radiotherapy phases and generated a phase-specific clinical, imaging or
quantitative output. Cross-cutting utilities provided information retrieval,
workflow logistics, deterministic calculation or platform guidance that could
be used in more than one phase. Each capability was exposed through the common
function registry and shared structured-output contract described above.
Supplementary Table~S1 provides a concise
description and headline evidence for all 26 functions; Supplementary
Table~S2 provides development and evaluation
data for functions with a formal task-level evaluation.

\noindent\textit{Diagnosis, Patient Evaluation \& Decision-Making.} Patient
evaluation uses the available documented clinical context, including
examination findings, diagnostic imaging, laboratory results and disease stage,
to inform treatment selection. The two implemented phase-specific functions
are RT-indication assessment and modality selection. RT-indication assessment
retrieves and cites disease-specific guidance from a locally curated guideline
corpus and returns an insufficient-evidence response when the retrieved
material does not support the request. Modality selection produces a
comparative assessment from the clinical context using the previously
evaluated fine-tuned language-model approach~\cite{wang2025finetuning}. When candidate-plan dose
information and a toxicity endpoint with verified coefficients are available,
it additionally performs a deterministic Lyman--Kutcher--Burman normal-tissue
complication-probability calculation; unsupported endpoints are not assigned
a numerical estimate~\cite{marks2010quantec}.

\noindent\textit{Simulation \& Planning.} Planning-directive generation
drafts a structured simulation and planning instruction from the diagnosis,
treatment intent, prescription and retrieved guidance~\cite{cao2025ctsim}.
MR-to-CT synthesis generates a synthetic CT from planning MRI and a body mask
for MR-only simulation~\cite{villegas2024syntheticct}. Deformable registration
produces a deformation field and propagated structures from baseline and
follow-up image data~\cite{ding2023dir}.
Auto-contouring returns target and OAR masks from planning CT and plan-context
information~\cite{zhang2024promptableseg}. Proton
dose prediction produces a three-dimensional PBSPT dose distribution from CT,
structures, beam information and prescription~\cite{zhang2023beammask,
zhang2026noisyprobing}. Plan optimization iteratively updates planning
objectives in response to treatment-planning-system feedback and returns the
resulting plan state~\cite{wei2025feasibility,wei2026dualagent}. For IMRT, this functionality
additionally uses multimodal LLM analysis of CT images annotated with target
and organ-at-risk contours to select beam templates before iterative plan
optimization~\cite{wei2026dualagent}.

\noindent\textit{On-Treatment Adaptive Management.} Verification CT synthesis
generates a CT-like image from on-treatment CBCT or orthogonal kV images for
adaptive workflow use~\cite{ding2024kvct}. Replan assessment combines
registered baseline and on-treatment anatomy with available dose information
to report deformation- and dose-based change metrics and, when configured,
a replan-trigger classification~\cite{feng2025oapt}. This output is presented
as decision support
and does not initiate replanning or replace clinician review.

\noindent\textit{Follow-up \& Outcomes.} Seven functions use a common
EHR-grounded reasoning mechanism to draft a treatment-completion note and to
assess recurrence, mandibular osteoradionecrosis, brain necrosis, cardiac
injury, radiation esophagitis and radiation pneumonitis from relevant
longitudinal clinical, imaging and dose records. Published task-specific
evaluations support the recurrence and mandibular-ORN components
~\cite{holmes2025outcomes}; the retrieved record categories and
condition-specific criteria differ by function. No outcome function
autonomously establishes a diagnosis.

\subsubsection{Cross-cutting utilities}
The nine utilities provide patient-status, implant, diagnostic-imaging and
In-Basket-message retrieval; clinical-trial search; schedule summary and
deterministic conflict checking; literature search; proton stopping-power
lookup; and platform guidance. Retrieval utilities return the corresponding
source-system records without reinterpretation. Trial search returns candidate
trials and their stored criteria but does not independently adjudicate
eligibility. Schedule checking identifies time overlaps deterministically;
proton stopping-power lookup uses locally stored reference tables; literature
search calls the NCBI E-utilities service; and platform guidance returns
registered static information. These utilities contribute workflow integration
rather than new predictive models and were evaluated by system-level function
selection, argument validity, dispatch and structured-return validity.

\subsection{Architecture-level evaluation}
\label{sec:methods-architecture-evaluation}
A reference set of 2,600 unique single-function requests was constructed to
cover all 26 registered capabilities, with 100 requests per function. For each
function, the set comprised 30 direct and explicit requests, 30 natural
clinical requests containing patient context, 20 requests using abbreviations,
paraphrases or reordered wording, and 20 deliberately challenging requests
designed to distinguish the target from a neighbouring function. Each request
had a prespecified intended function, and schema conformance was checked
automatically against the function registry. Every request was submitted in
three separate, newly initialized conversation sessions using the same input
and system configuration, with the conversational state reset before each
run, yielding 7,800 executions. The 2,600 unique requests were the test units;
the three executions of each request were treated as repeated observations
for evaluating run-to-run reproducibility.

Function-selection accuracy was defined as exact agreement between the
selected and annotated functions. Argument validity required the generated
arguments to satisfy the schema of the function actually selected by the
controller; this endpoint was therefore evaluated independently of whether
that function matched the annotated target. Technical dispatch success
required the generated call to reach a registered backend without a routing
or transport error, and structured-output success required the returned
response to be parseable under the shared result contract. These four
endpoints were analysed as marginal proportions and were not combined into an
end-to-end success rate. Repeated-run performance was summarized at the level
of the 2,600 unique requests as the number of sessions, from zero to three, in
which the intended function was selected.

A distinct missing-input set comprised 520 unique requests, with 20 requests
per registered function, each repeated in three newly initialized,
context-reset sessions for 1,560 executions. We recorded separately whether
the controller detected incomplete required information and whether it issued
a targeted clarification or withholding response without an execution attempt.
An attempt to call a backend despite unresolved required input was counted as
an inappropriate execution attempt.
Orchestration overhead was measured from request receipt to backend dispatch,
excluding specialist-model inference and external-service response time, and
is reported as the median and interquartile range.

Two investigators familiar with radiation-oncology workflows and the function
registry constructed the benchmark by creating clinically grounded seed
templates for each function.
Candidate benchmark requests, including direct, clinical-context, abbreviated,
paraphrased and reordered variants, were generated from these templates using
a language model distinct from the evaluated controller. Investigators
reviewed and edited each item to preserve a single intended clinical task
without explicitly naming the target function.
For each function, one to three neighbouring functions were prespecified on
the basis of overlapping scope, inputs or outputs. Two reviewers independently
assigned the intended-function label to every item, with disagreements
resolved by a third reviewer before evaluation. The benchmark was finalized
after the controller prompt and registry were locked and was not used for
prompt refinement, registry modification or backend tuning.

Missing-input requests were generated by withholding one schema-required input
and manually confirming that execution should require clarification. Twenty
functions were exercised through registered-service interfaces and six
computationally intensive imaging or planning functions through replay or test
interfaces, corresponding to 6,000 and 1,800 normal-request executions,
respectively. Replay and test backends reproduced the registered input and
output contracts to isolate orchestration from repeated specialist-model
inference. Controller overhead was calculated across the 7,719 executions
that successfully reached backend dispatch, from request receipt to issuance
of the validated backend request; specialist-model inference, asynchronous
jobs and external-service response time were excluded.

\subsection{Comparative architecture-ablation evaluation}
\label{sec:methods-architecture-ablation}
We prespecified comparative ablations to identify the contributions of
language-based routing, persistent workflow state and deterministic validation
to the system-level results. All configurations used the same locked function
registry, benchmark inputs, controller decoding settings and scoring rules as
the corresponding primary evaluation. The complete RadOnc-Agent configuration
served as the reference. Three comparators were evaluated only on endpoints to
which they could meaningfully contribute.
First, an embedding-based router was evaluated on the 2,600 unique
single-function requests. The router selected the registered function whose
standardized description had the highest cosine similarity to the request,
using the same fixed text-embedding model and registered function description
for every request. It did not generate function arguments or dispatch a
backend; its sole endpoint was exact intended-function selection, overall and
by request type. Second, a stateless-controller ablation used the same language
model, registry and deterministic validation layer as the complete system, but
discarded the patient/workflow state and preceding structured output after each
turn. The next turn therefore contained only the newly released scenario input.
This configuration was evaluated on the 200 cross-stage scenarios, each run
three times in independently initialized sessions, using completion,
trajectory adherence, information transfer, context retention and unnecessary
calls as endpoints.

Third, a validation-gate ablation used the same controller and stateful
execution logic but bypassed the pre-dispatch schema and patient/course/plan
identifier checks. It was evaluated exclusively through replay or test
interfaces, and no bypassed call was sent to an actual registered service. The
missing-input benchmark was used to compare controller attempts to proceed
with the resulting unsafe backend dispatches. A prespecified
identifier-integrity subset was constructed by substituting an inconsistent
patient, course or plan identifier at eligible replay/test hand-offs and
comprised 360 executions. Its endpoints were detection of the mismatch, safe
rejection and mismatched dispatch. Each comparator used the same repeated-run
design as its corresponding reference benchmark. Benchmark items were not used
for comparator selection or tuning.

\subsection{Cross-stage workflow evaluation}
\label{sec:methods-cross-stage-evaluation}
Multi-stage scenarios were organized into four workflow families representing
decision-to-planning, planning-to-adaptation, treatment-to-follow-up and
complete phase-1-to-phase-4 execution (Table~\ref{tab:cross-stage-workflows}).
The evaluation comprised 200 unique scenarios, 50 per family. Each scenario
was executed three times in a separate, newly initialized session after
clearing the conversational state, yielding 600 workflow executions. The
scenarios were the unique test units, and repeated executions were clustered
within scenario. A prespecified robustness subset of 90 executions received
either a missing-input or transient-backend perturbation; all other executions
used the original scenario inputs and locked system settings. Of the 600
executions, 360 used registered-service execution and 240 used replay or test
execution, as defined in the system-implementation section. Replay or test
mode isolated orchestration behaviour from repeated specialist-model
inference.
Cross-cutting utilities were invoked when required but were not treated as an
additional clinical phase.

Each scenario comprised a scripted sequence of clinical inputs and user
objectives expressed without naming the functions to be called. Baseline
patient information initiated the session, and additional imaging, treatment
or follow-up artefacts were released at prespecified turns when required by
the workflow family. The agent was provided only with these scenario inputs;
the adjudication record was withheld and specified the required function
calls, partial-order constraints, required data dependencies between calls and
acceptable optional calls. Figure~\ref{fig:e2e}a shows an illustrative
synthetic full-pathway scenario in this format. A workflow was considered
complete only if all required calls returned a parseable output or an
explicitly accepted non-execution outcome in a prespecified missing-input
scenario, such as a clarification request without backend dispatch.

Trajectory adherence required completion of every required function in an
order consistent with the reference constraints. Information-transfer
accuracy was scored at each eligible hand-off by comparing the patient
identifier and required structured fields received by the downstream function
with the upstream result and scenario record. Context retention required the
agent to preserve the active patient, treatment intent, prescription and
current workflow stage across turns without importing information from another
case. The 600 executions contained 3,150 required calls, 2,550 eligible
interfunction hand-offs and 4,800 turns at which context retention was scored.
Calls not included among the required or acceptable optional functions were
counted as unnecessary; the excess-call ratio used the 3,150 required calls as
its denominator. Recovery was evaluated only in the prespecified 90-execution
robustness subset and required subsequent successful workflow completion after
clarification or restoration of the backend.
Backend execution mode was logged for each call as registered-service, replay or
test so that orchestration-only results were not presented as actual-service
integration or as evidence of routine clinical availability.

Two investigators familiar with radiation-oncology workflows and the
registered functions constructed the scenarios. Two additional reviewers
independently assessed clinical plausibility, uniqueness of the intended
trajectory and adequacy of the reference calls; disagreements about required
functions, permissible ordering or acceptable optional calls were resolved by
consensus before evaluation. Scenario construction was completed after the
controller prompt, registry and backend interfaces were locked, and no
evaluation scenario was used for prompt refinement or registry modification.
Scenario difficulty was balanced prospectively by the number of clinical
stages, required calls, interfunction dependencies and conversational turns.
Across the 150 executions in each family, the three shorter families contained
690--720 required calls, 540--570 eligible hand-offs and 1,050 scored turns;
full-pathway workflows contained 1,020 required calls, 870 hand-offs and
1,650 scored turns by design. Within each family, 90 executions used
registered-service execution and 60 used replay or test execution. The 90-execution robustness
subset comprised 45 missing-input and 45 transient-backend perturbations,
distributed approximately evenly across workflow families. Missing-input
recovery required targeted clarification followed by resumption of the
reference trajectory after the information was supplied; transient-failure
recovery required completion after service restoration.
Cross-case context isolation was assessed by interleaving sessions for
distinct synthetic patients and checking the active patient identifier in the
session state and dispatched calls at each scored turn.

\subsection{Real-patient cross-stage workflow-execution evaluation}
The real-patient evaluation comprised 60 de-identified patient records, each
contributing a decision-to-planning and a planning-to-adaptation workflow. The
controller configuration, function registry and specialist backends were held
fixed throughout the evaluation. All required functions were executed through
their registered services in a retrospective offline environment, and no
output was used to direct patient care.
For decision-to-planning workflows, the input package contained recorded
diagnosis and course metadata, treatment intent and prescription, and the
planning CT. The reference trajectory required planning-directive generation
followed by auto-contouring. For planning-to-adaptation workflows, the input
package contained the planning CT, contours, reference dose and at least one
verification CT. The reference trajectory required deformable registration
followed by replan assessment. Reference trajectories and required function
arguments were specified before execution from the registered function
contracts and available source objects. Decision-to-planning inputs included
the patient and course identifiers, disease site and stage when available,
treatment modality, prescription, planning-CT series and image volume.
Planning-directive generation used the registered clinical-text service and
auto-contouring used the registered segmentation service. Planning-to-adaptation
inputs additionally included the reference-plan identifier, structure set,
dose distribution, verification-CT series and image volume, and the
associated treatment fraction or date; registration and replan assessment used
their registered services. When more than one verification CT was available, we
selected the earliest scan acquired after treatment initiation that met the
prespecified technical eligibility criteria and could be linked unambiguously
to the reference plan. For multiple eligible scans on the same date, the scan
closest in time to the documented treatment fraction was selected.

A recorded subsequent-plan-change label was derived from the linked treatment
record when a later plan was present in the same treatment course. This label
was used only to evaluate detection of a subsequently recorded plan change;
it was not used as a reference for the clinical appropriateness of adaptive
replanning. A plan change required a new treatment-plan identifier after the
selected verification CT, within the same course, that replaced the reference
plan for one or more subsequently delivered fractions and was associated with
a new planning or dose object. Preplanned sequential, boost or cone-down
plans, field-in-field or delivery-only revisions, and plans documented before
the selected verification CT were not counted as adaptive plan changes. When
multiple qualifying later plans were present, the first was used for the binary
label. Cases with ambiguous linkage between a later plan and the reference
course were excluded from the plan-change comparison but retained in the
technical workflow-execution analysis.

Each workflow was executed three times with identical inputs and locked system
settings. Every execution began in a separately initialized, context-reset
conversation session. The primary outcomes were required-function completion,
trajectory adherence, argument validity, patient/course/plan identifier
propagation, structured-output validity and backend-dispatch success.
Secondary outcomes were context retention, unnecessary calls, recovery after
injected missing inputs or transient backend failures, controller overhead and
run-to-run reproducibility. Evaluation was performed against the prespecified
technical reference trajectory and execution logs.
Identifier propagation was evaluated only for eligible hand-offs, defined as
calls with an available upstream output and a downstream function contract
requiring a patient, course or plan identifier. For the one-per-workflow
comparison with subsequently recorded plan-change status, the output from the
first prespecified clean execution was used; the other two clean executions
were reserved for the reproducibility analysis.
The three clean executions per workflow comprised the primary evaluation
(360 executions). A separate prespecified stress test comprised 48 additional
executions, including 24 with a missing required input and 24 with a transient backend
failure. Recovery required the agent to resolve the injected condition and
complete the prespecified trajectory; these perturbed executions were not
included in the clean-execution completion, trajectory or reproducibility
endpoints.

\subsection{Task-specific function-level evaluation}
Function-level evaluations used the task-appropriate unit, reference standard
and native endpoint for each callable capability; heterogeneous endpoints were
not pooled. For capabilities evaluated in previously reported task-specific
studies, the original held-out cohorts, reference standards and analysis
definitions were retained~\cite{wang2025finetuning,cao2025ctsim,
zhang2023beammask,
zhang2026noisyprobing,wei2025feasibility,wei2026dualagent,ding2024kvct,
holmes2025outcomes}. Internal task-specific evaluations without a public
report are described in this study and identified as such in the Supplementary
Information. Function descriptions and resulting estimates are
summarized in Supplementary Table~S1.
Evaluations conducted specifically for the current framework are described
below; development data, evaluation data and reference standards are reported
in Supplementary Table~S2.

Guideline-grounded RT-indication assessment was evaluated with 300
prespecified questions across ten disease sites, comprising 240 questions
supported by the indexed guideline corpus and 60 deliberately unsupported or
out-of-corpus questions. Clinically grounded questions were balanced across
disease sites and finalized before system evaluation. Two radiation oncologists
independently established the reference answer, supporting passage and
expected refusal status while blinded to agent outputs; disagreements were
resolved by consensus and third-reviewer adjudication when needed.

MR-to-CT synthesis used the publicly
released 578 paired MRI--CT cases from the SynthRAD2025 MRI-to-CT dataset,
including 221 head-and-neck, 182 thoracic and 175 abdominal cases~\cite{synthrad2025}.
Cases were split at the patient level with stratification by anatomical region
into 462 training, 58 validation and 58 internal held-out test cases. Images
were resampled, intensity-normalized and cropped or padded using the
training-set pipeline; the validation MAE was used for checkpoint selection.
MAE and SSIM were calculated within the reference body contour.

Replan assessment was evaluated within the real-patient planning-to-adaptation
workflow evaluation using the same 60 Mayo Clinic Arizona patients described
above. The function output from the first prespecified clean execution was
compared with the recorded subsequent-plan-change endpoint defined above.
Completion-note generation was evaluated in 200 completed radiotherapy courses.
Two radiation oncologists
independently compared prespecified diagnosis, treatment, toxicity and
follow-up fields with the source record while blinded to one another's ratings.
A clinically important omission was one capable of materially changing
interpretation of delivered treatment, toxicity or planned follow-up;
substantive edits added, removed or changed clinically meaningful factual
content. Disagreements were resolved by consensus.

Brain-necrosis assessment was evaluated in 200 patients, including 60
adjudicated positive cases. Brain necrosis was established from
post-radiotherapy imaging, treatment history and clinical documentation.
Cardiac injury, radiation esophagitis and radiation pneumonitis assessment
were evaluated separately in the same 438-patient cohort (275 with esophageal
cancer and 163 with lung cancer), with one patient-level assessment per
endpoint. These workflows used GPT-4.1. Cardiac-injury assessment first
summarized longitudinal cardiac evidence and then assessed event presence,
severity grade and event types. Esophagitis and pneumonitis assessment
retrieved relevant original clinical notes and reports before determining event
presence and grade; no intermediate note summarization was used. Clinician-
adjudicated reference standards were used for event status and grade. Cardiac
event-type matching was evaluated separately in 165 reference-positive cases
with evaluable event-type information. Patient status, implant retrieval, diagnostic imaging, In-Basket
messages, trial search, schedule/conflict checking, platform guidance, PubMed
search and PSTAR lookup were treated as direct
retrieval, deterministic-calculation, static-response or external-API
utilities. They were assessed through system-level function selection,
argument validity, dispatch and structured-return validity rather than a
separate predictive-performance study.

\subsection{Evaluation metrics}
System-level orchestration endpoints were defined in the corresponding
evaluation subsections above. For guideline assessment, top-5 evidence
retrieval was successful when at least one adjudicated supporting passage was
returned among the first five results. Answer accuracy required agreement with
the adjudicated answer or, for an unsupported question, an appropriate
refusal. Citation correctness required the cited passage to support the
generated statement.
Binary clinical endpoints were summarized using sensitivity, specificity,
precision, recall, F1 score and accuracy relative to the prespecified
reference standard. The positive class was the adjudicated condition for
outcome assessments and a subsequently recorded plan change for replan
assessment. AUROC was calculated from a continuous model score before applying
the prespecified operating threshold. Ordinal toxicity grades were evaluated
using exact agreement, macro-F1 and weighted Cohen's $\kappa$. Refusal rates
used unsupported or out-of-corpus questions as the denominator.

Deformable registration and auto-contouring were evaluated using Dice
similarity relative to expert manual contours; inference time was additionally
reported for auto-contouring.
MR-to-CT synthesis was evaluated using body-region Hounsfield-unit mean
absolute error (MAE) and structural similarity relative to the reference CT.
Verification CT synthesis was evaluated using CT MAE, CT-number error,
dose-recalculation gamma passing rate, setup-shift error and reconstruction
time. Dose prediction used target, OAR and body gamma passing rates and
inference time relative to the reference dose. Gamma criteria and dose
thresholds are reported with each relevant experiment. Factual
accuracy for structured summaries was calculated at the prespecified field
level; completeness was the proportion of required fields present, and a
clinically important omission was an absent item judged capable of changing
clinical interpretation or follow-up. Usability ratings distinguished outputs
requiring no modification, minor modification, major modification or rejection
according to the definitions used in the system-level evaluation.

\subsection{Statistical analysis}
The analysis unit was the unique request, scenario, workflow, patient, imaging
episode or retrieved item, as specified for each experiment. Repeated
executions were not treated as independent test units and were summarized or
analysed as observations clustered within the originating request, scenario or
workflow. Categorical outcomes are reported as counts and percentages;
continuous outcomes are reported as mean$\pm$s.d. or median and interquartile
range according to their distribution. No imputation was performed. Each
endpoint used its prespecified evaluable denominator; technical failures were
counted as failures for completion-related endpoints, whereas cases lacking
objects required for a specific endpoint were excluded only from that endpoint.

Workflow-family, backend-mode, failure-category and reproducibility analyses
were prespecified descriptive secondary analyses. For the real-patient
evaluation, repeated executions were clustered within workflow and workflows
were clustered within patient. Benchmark request and scenario counts were
prespecified to balance registered functions, request types and workflow
families. The 60-patient cohort provided balanced coverage of five anatomical
groups (12 patients per group) and both real-patient workflow families; three
clean executions assessed run-to-run reproducibility, and 48 perturbed
executions balanced missing-input and transient-backend stress testing. No
formal power calculation was performed because endpoints were descriptive
technical measures rather than clinical-effect comparisons.

For comparative ablations, the complete configuration and the relevant
comparator were evaluated on the same unique requests or scenarios. We report
the paired count and percentage for each configuration and the absolute paired
difference in percentage points. The embedding-router comparison was limited
to function selection; the stateless-controller comparison was limited to
cross-stage execution endpoints; and the validation-gate comparison was
limited to replay or test safety endpoints.

Unless otherwise specified, confidence intervals were two-sided 95\%
percentile intervals from 10,000 non-parametric bootstrap resamples generated
with the PCG64 random-number generator (seed 20261011). System-level
endpoints were resampled at the level of the unique request (including the
missing-input benchmark) or unique scenario, with all repeated executions
retained within the sampled cluster. The routing, state and validation
comparisons used paired request- or scenario-level resampling, as appropriate,
and report percentile intervals for the absolute paired difference in
percentage points. Real-patient workflow endpoints used patient-level cluster
resampling, retaining each patient's workflows and repeated runs. Binary
patient-level endpoints used Wilson 95\% confidence intervals; F1 and weighted
Cohen's $\kappa$ used patient-level bootstrap resampling; AUROC used
patient-level stratified bootstrap resampling; and continuous patient-level
mean$\pm$s.d. endpoints used $t$-distribution confidence intervals. For
zero-event clustered endpoints, two-sided Wilson 95\% confidence intervals
were calculated at the corresponding cluster level. For
image-derived endpoints, the resampling unit followed the patient or
hierarchical patient-level design specified for the individual task. Exact point
estimates and 95\% confidence intervals are reported in Supplementary
Tables~S3, S4 and S5.

%% =======================================================
%% Unnumbered sections (Nature-family style)
%% =======================================================

\section*{Data Availability}
The de-identified Mayo Clinic Arizona clinical imaging, treatment-planning and
longitudinal record data used for real-patient and task-specific evaluations
cannot be publicly shared because they remain subject to institutional review
board and institutional data-governance restrictions. The specialist functions
integrated in RadOnc-Agent were developed and evaluated in task-specific
studies using combinations of institutional and public data; the public
resources listed below are components of those original development or external
evaluation datasets, rather than the sole data used to develop the models.
The MRI-to-CT synthesis study used the
\href{https://doi.org/10.5281/zenodo.14918089}{SynthRAD2025 collection}.
The deformable-registration study additionally included the public
\href{https://www.cancerimagingarchive.net/collection/4d-lung/}{4D-Lung},
\href{https://fdat.uni-tuebingen.de/records/75kj1-64747}{LongCT-Lung},
\href{https://www.cancerimagingarchive.net/analysis-result/nlst-new-lesion-longct/}{NLST-New-lesion-LongCT},
\href{https://www.cancerimagingarchive.net/collection/pancreatic-ct-cbct-seg/}{Pancreas-CT}
and
\href{https://www.cancerimagingarchive.net/collection/rider-lung-ct/}{RIDER Lung CT}
collections as external cohorts. The auto-contouring study additionally used
the public
\href{https://zhaoziheng.github.io/SAT/}{SAT-DS corpus} for image-encoder
pretraining, and the \href{https://github.com/ababier/open-kbp}{OpenKBP},
\href{https://www.cancerimagingarchive.net/collection/lctsc/}{LCTSC} and
\href{https://www.cancerimagingarchive.net/collection/soft-tissue-sarcoma/}{TCIA
Soft-Tissue-Sarcoma collection} for external evaluation. The latter contains
51 soft-tissue sarcoma cases with tumour (GTV) contours, rather than
radiotherapy CTV contours, and no OAR annotations. The constituent datasets
in SAT-DS and the Soft-Tissue-Sarcoma collection are available under the terms
specified by their original providers. The NIST PSTAR reference tables used by the
stopping-power lookup utility are available from the
\href{https://physics.nist.gov/PhysRefData/Star/Text/PSTAR.html}{NIST PSTAR
database}. Supplementary Table~S2 identifies
the data source for each task-specific function.

\section*{Code Availability}
The RadOnc-Agent source code and institutionally trained model weights are not
publicly available because the framework is maintained within the Mayo Clinic
enterprise environment and includes clinical-system integrations, security
configurations and models governed by institutional privacy and data-use
requirements. All 26 functions were implemented and integrated; text-only
functions were available to authorized clinical users, whereas some
image-processing functions remained restricted to retrospective research
because clinical image read or write permissions had not been enabled. On
request from editors or reviewers, the authors will seek institutional approval
to provide selected non-sensitive materials under controlled conditions,
potentially including synthetic benchmark inputs, intended-function labels and
reference trajectories, scoring and aggregate-analysis code, and a redacted
machine-readable function-contract schema. Patient data, production-system
routes or credentials, enterprise integration code and institutionally trained
model weights will not be shared. The Article and Supplementary Information
provide the function descriptions, evaluation protocols and aggregate results
needed to interpret the study.

\section*{Funding}
This work was supported by NIH/NIBIB R01EB293388, NIH/NCI R01CA280134 and
Schmidt Sciences.

\section*{Author contributions statement}
C.J., S.W., S.Z., J.L. and J.C. contributed to the development and
implementation of specialist functions integrated in RadOnc-Agent. C.J.,
S.W. and S.Z. drafted the manuscript. W.L. supervised the study. All authors
contributed to interpretation of the results, critically reviewed the
manuscript and approved the final version.

\section*{Competing interests}
The authors declare no competing interests.

%% =======================================================

\bibliography{radonc_agent_paper}
\onecolumn
\captionsetup{font=scriptsize,skip=3pt}
\section*{Tables}
\begin{center}
\fontsize{7.6}{8.6}\selectfont
\setlength{\tabcolsep}{3pt}
\renewcommand{\arraystretch}{1.14}
\captionof{table}{\textbf{Workflow-scope comparison of representative medical
and radiotherapy agent systems.} The comparison describes the scope and
evaluation reported for each system and is not a head-to-head comparison of
language-model or task-specific performance. Context continuity refers to the
clinical or computational state explicitly maintained within the reported
workflow.}
\label{tab:prior-agent-scope}
\begin{tabular}{>{\raggedright\arraybackslash}p{2.10cm}
                >{\raggedright\arraybackslash}p{2.30cm}
                >{\raggedright\arraybackslash}p{2.70cm}
                >{\raggedright\arraybackslash}p{2.40cm}
                >{\raggedright\arraybackslash}p{2.30cm}
                >{\raggedright\arraybackslash}p{3.20cm}}
\toprule
\rowcolor{tablehead}
\textbf{System} & \textbf{Clinical scope} & \textbf{Integrated capabilities} &
\textbf{Cross-stage composition} & \textbf{Context continuity} &
\textbf{Patient-data evaluation} \\
\midrule
MIRA~\cite{mira2026}
& General medicine; emergency-care episode
& History taking, laboratory, imaging and microbiology tests, diagnosis,
treatment and disposition
& Diagnostic and therapeutic actions within one emergency-care episode
& Episode-level state in a sandboxed EHR
& 574 retrospective MIMIC-IV cases evaluated in sandboxed simulated workflows
\\
\midrule
\rowcolor{tablesubtotal}
DOLA~\cite{nusrat2025dola}
& Radiotherapy inverse planning
& Local language model, retrieval, reinforcement learning and commercial-TPS
optimization
& No; confined to one planning task
& Iterative optimization state
& 18 retrospective prostate-cancer cases \\
\midrule
Yang et al.~\cite{yang2026referencefree}
& Radiotherapy inverse planning
& Clinical-objective interpretation, plan evaluation and commercial-TPS
optimization
& No; confined to one planning task
& Iterative planning and optimization state
& 20 retrospective head-and-neck cancer cases \\
\midrule
\rowcolor{tablesubtotal}
Compound-AI planning platform~\cite{maniscalco2026compound}
& Radiotherapy treatment planning
& Seven-agent orchestration, directive-conditioned dose prediction,
institutional knowledge retrieval and commercial-TPS optimization
& No; confined to one planning episode
& Iterative dose, objective and TPS state
& 60 retrospective brain, lung and prostate cases \\
\midrule
Knowledge-guided M2M agent~\cite{abrar2026m2m}
& Radiotherapy treatment planning
& LLM-based iterative planning guided by treatment-planning-parameter knowledge
from reinforcement learning
& No; confined to one planning task
& Iterative treatment-planning-parameter state
& Three planning scenarios spanning standard and constrained prostate cases and
liver cases \\
\midrule
\rowcolor{tablesubtotal}
One-for-All adaptive-planning agent~\cite{arpa2026}
& Daily online adaptive radiotherapy
& Synthetic CT, multimodal alignment, segmentation and adaptive plan generation
& Within an adaptive-planning episode; not across the broader care pathway
& Daily-fraction and adaptive-episode context
& Multiple head-and-neck, lung, abdominal and prostate datasets with photon and
proton therapy \\
\midrule
\rowcolor{tablesection}
\textbf{RadOnc-Agent (this study)}
& Four-phase radiotherapy care pathway
& 17 phase-specific functions and 9 cross-cutting clinical utilities
& Across diagnosis, patient evaluation, decision-making, planning, adaptation
and follow-up
& Patient-, course- and plan-level context across stages
& 60 patients and 120 retrospective workflow instances; complete four-phase
workflows evaluated synthetically \\
\bottomrule
\end{tabular}
\end{center}
\clearpage
\begin{center}
\fontsize{8.2}{9.4}\selectfont
\setlength{\tabcolsep}{4pt}
\renewcommand{\arraystretch}{1.18}
\captionof{table}{\textbf{Cross-stage workflow families used to evaluate
longitudinal coordination by RadOnc-Agent.}}
\label{tab:cross-stage-workflows}
\begin{tabular}{>{\raggedright\arraybackslash}p{2.45cm}
                >{\raggedright\arraybackslash}p{1.70cm}
                >{\raggedright\arraybackslash}p{5.40cm}
                >{\raggedright\arraybackslash}p{4.90cm}}
\toprule
\rowcolor{tablehead}
\textbf{Workflow family} & \textbf{Phase transition} &
\textbf{Representative call trajectory} & \textbf{Primary evaluation objective} \\
\midrule
Decision-to-planning
 & Phase 1 $\rightarrow$ Phase 2
 & RT-indication assessment or modality selection $\rightarrow$ planning-directive generation $\rightarrow$ auto-contouring $\rightarrow$ dose prediction or plan optimization, as specified
 & Transfer of treatment intent, modality and prescription into the simulation-and-planning context \\
\midrule
\rowcolor{tablesubtotal}
Planning-to-adaptation
 & Phase 2 $\rightarrow$ Phase 3
 & Planning CT, contours and dose $\rightarrow$ verification CT synthesis, when required $\rightarrow$ deformable registration $\rightarrow$ replan assessment
 & Transfer of the reference plan and anatomy into on-treatment comparison and adaptive assessment \\
\midrule
Treatment-to-follow-up
 & Phase 3 $\rightarrow$ Phase 4
 & Treatment-delivery and adaptation record $\rightarrow$ completion-note generation $\rightarrow$ recurrence or specified toxicity assessment
 & Retention of delivered-treatment history during longitudinal outcome assessment \\
\midrule
\rowcolor{tablesubtotal}
Full-pathway workflow
 & Phase 1 $\rightarrow$ Phase 2 $\rightarrow$ Phase 3 $\rightarrow$ Phase 4
 & A clinically appropriate subset of core functions across all four phases, supplemented by cross-cutting utilities when required
 & End-to-end function sequencing, data transfer, context continuity and completion \\
\bottomrule
\end{tabular}
\end{center}
\clearpage
\section*{Figures}
\begin{center}
\centering
\includegraphics[width=0.94\textwidth,height=0.84\textheight,keepaspectratio]{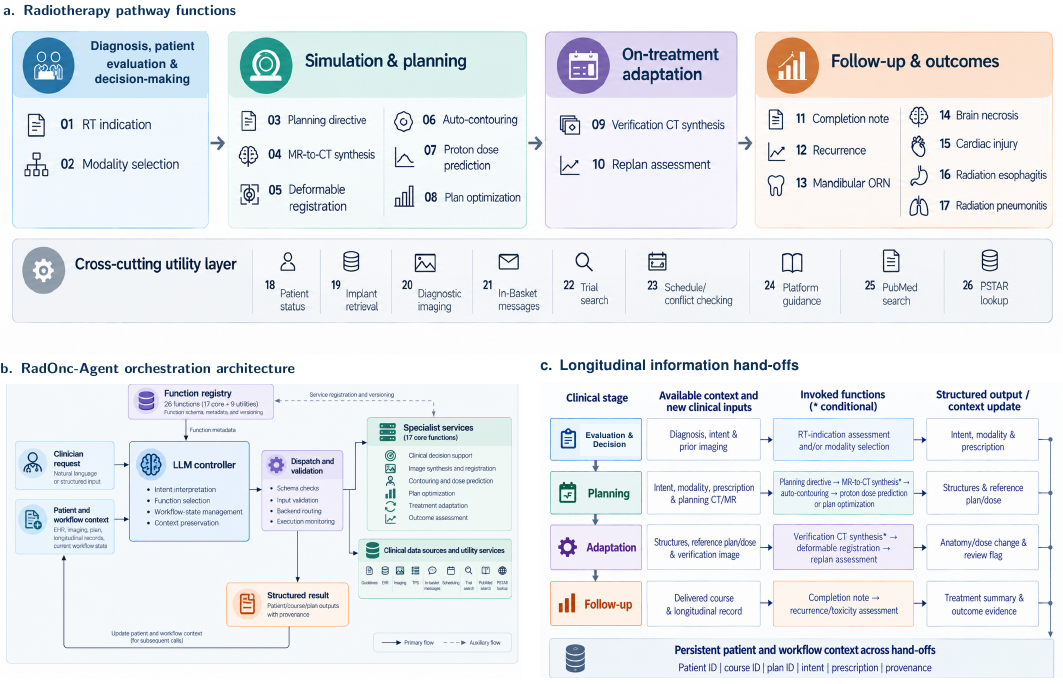}
\captionof{figure}{\textbf{Overview of RadOnc-Agent across the radiotherapy
pathway.} \textbf{a,} Seventeen core functions are organized across four
clinical phases, with nine cross-cutting utilities. \textbf{b,} A
language-model controller combines clinician requests and patient context,
selects and validates registered functions, dispatches calls to specialist
services or clinical data and utility services, and returns provenance-tracked
structured results. \textbf{c,} An illustrative workflow beginning with the
Diagnosis, Patient Evaluation \& Decision-Making phase (abbreviated
``Evaluation \& Decision'' in the panel) shows how stage-specific inputs and
retained context are routed to registered functions
and returned as structured context updates across four clinical stages.
Patient-, course- and plan-level identifiers and provenance are preserved
across hand-offs; asterisks denote conditionally invoked functions.}
\label{fig:overview}
\end{center}
\clearpage

\begin{center}
\centering
\includegraphics[width=0.94\textwidth,height=0.84\textheight,keepaspectratio]{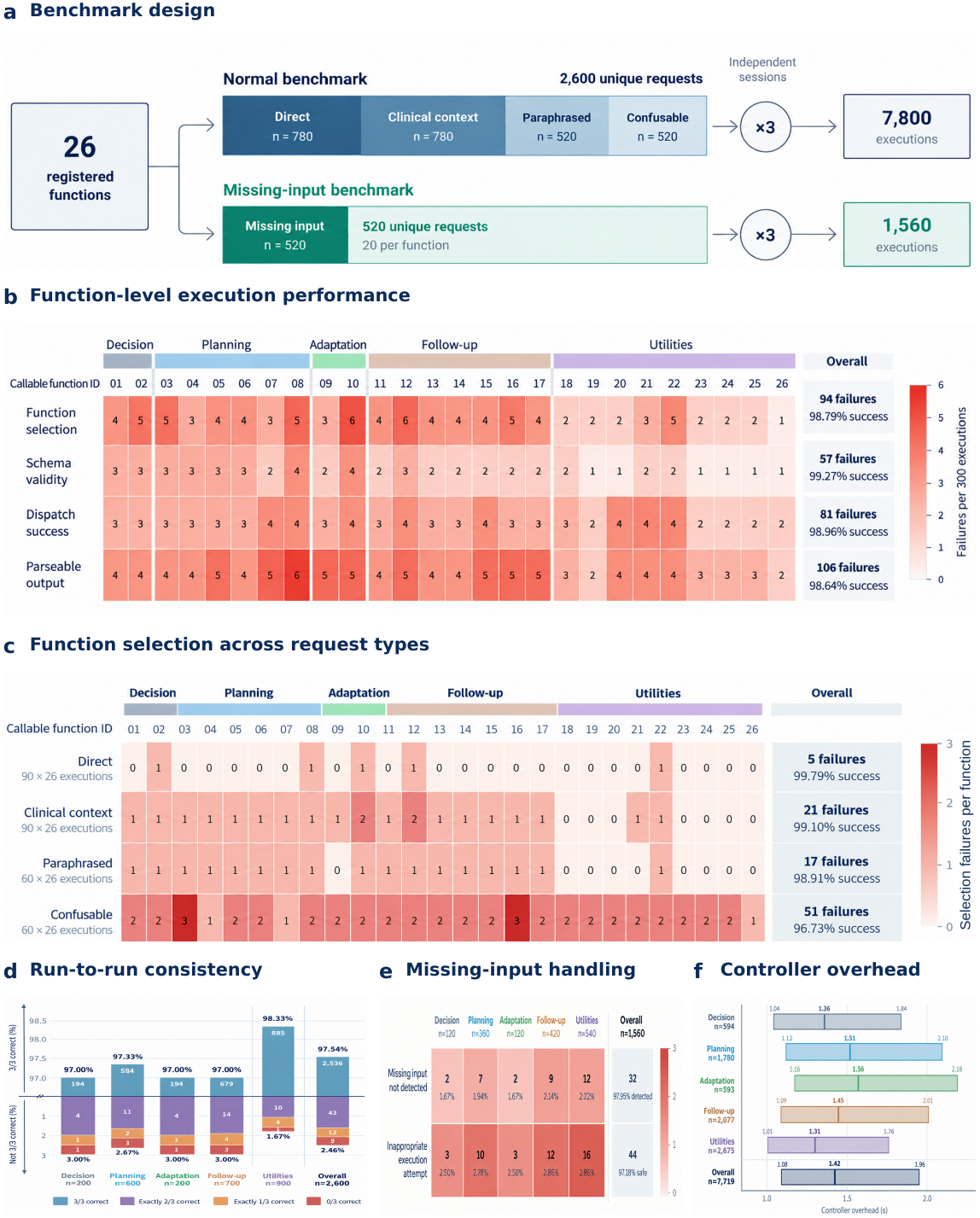}
\captionof{figure}{\textbf{Function-orchestration performance of
RadOnc-Agent.} \textbf{a,} Two parallel benchmarks used normal
single-function requests (2,600 unique requests; 100 per function, comprising 30 direct,
30 clinical-context, 20 paraphrased or reordered and 20 confusable) and
missing-input requests (520 unique requests; 20 per function). Each request
was evaluated in three independent context-reset sessions, yielding 7,800 and
1,560 executions, respectively. \textbf{b,} Function-level failures in
selection, schema validity, backend dispatch and parseable output across all
26 callable functions; cells show failures per 300 executions, and the right
column gives aggregate counts and success rates. \textbf{c,} Function-selection
failures by request type and callable function; the right column gives
aggregate counts and success rates for each request type. \textbf{d,}
Run-to-run consistency across the 2,600 unique requests, grouped by clinical
phase and utilities. Upper bars show requests correctly routed in all three
runs; lower stacked bars show requests correct in two, one or no runs.
\textbf{e,} Missing-input detection and safe-handling failures by function
group, with counts and rates shown in each cell. \textbf{f,} Median controller
orchestration overhead by function group; boxes span the interquartile range
and vertical lines indicate the median. Backend inference time is excluded.
Callable function identifiers in \textbf{b} and \textbf{c} correspond to
Supplementary Table~S1. Exact 95\% confidence intervals are reported in
Supplementary Table~S3.}
\label{fig:architecture-evaluation}
\end{center}
\clearpage

\begin{center}
\centering
\includegraphics[width=0.94\textwidth,height=0.84\textheight,keepaspectratio]{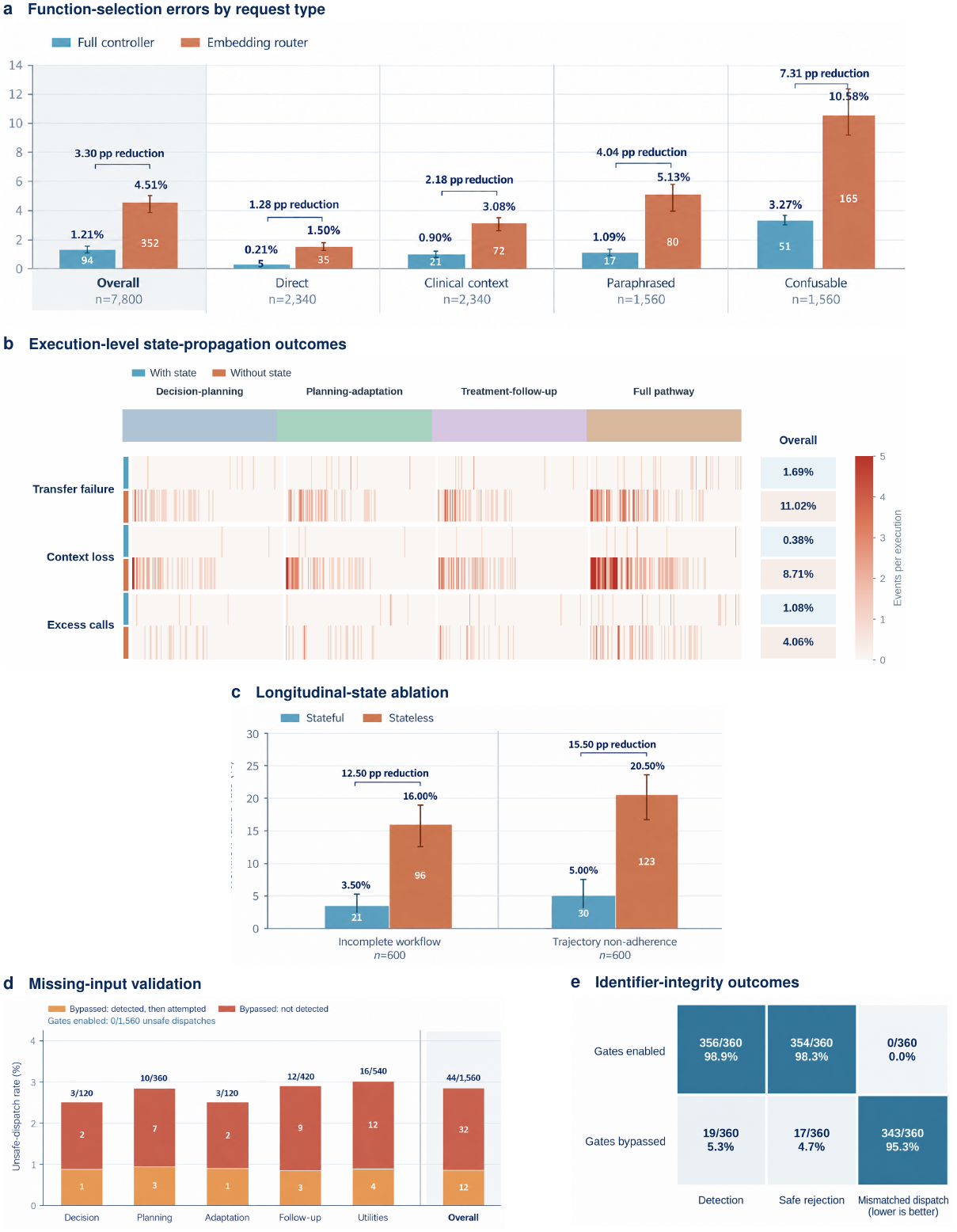}
\captionof{figure}{\textbf{Architecture ablations quantify the contributions
of language-based routing, persistent state and deterministic validation.}
\textbf{a,} Function-selection error rates for the full controller and an
embedding-based router in the 7,800-execution single-function benchmark,
overall and by request type. \textbf{b,} Execution-level information-transfer
failures, context losses and
excess calls with and without longitudinal state, arranged by workflow family;
overall event rates are shown at right. \textbf{c,} Incomplete-workflow and
trajectory-non-adherence rates for stateful and stateless controllers (600
executions). \textbf{d,} Unsafe dispatches after
validation gates were bypassed, grouped by function group and separated by
whether the missing input was detected before the execution attempt. Enabled
gates permitted no unsafe dispatch among 1,560 executions. \textbf{e,}
Identifier-mismatch detection, safe rejection and mismatched dispatch with
validation gates enabled or bypassed (360 replay/test executions per
condition); cell shading represents the event rate. Labelled differences in
\textbf{a} and \textbf{c} are percentage points. Exact 95\% confidence
intervals are reported in Supplementary Table~S4.}
\label{fig:architecture-ablation}
\end{center}
\clearpage

\begin{center}
\centering
\includegraphics[width=0.94\textwidth,height=0.84\textheight,keepaspectratio]{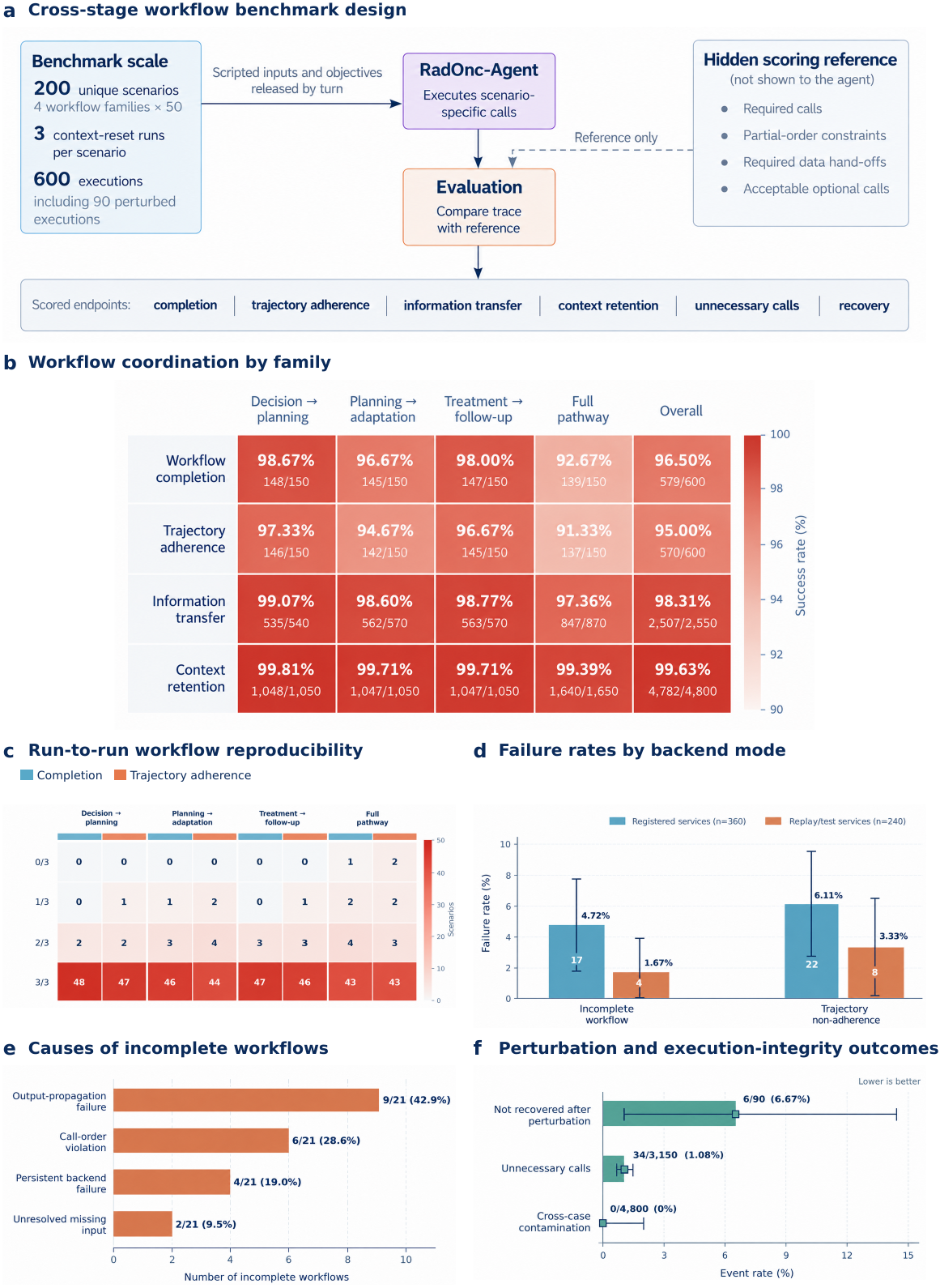}
\captionof{figure}{\textbf{Cross-stage workflow coordination by
RadOnc-Agent.} \textbf{a,} Synthetic benchmark and scoring design. The 200
unique scenarios comprised 50 decision-to-planning, planning-to-adaptation,
treatment-to-follow-up and full-pathway scenarios. Each was executed three
times in independently initialized, context-reset sessions (600 executions),
including a prespecified 90-execution perturbation subset. Scripted clinical
inputs and objectives were released by turn; the agent did not receive the
hidden reference calls, constraints or hand-offs used for scoring.
\textbf{b,} Workflow completion, trajectory adherence, information transfer
and context retention by workflow family and overall. \textbf{c,} Run-to-run
reproducibility across three executions of each scenario, shown as the number
of scenarios satisfying completion or trajectory adherence in zero to three
runs. \textbf{d,} Failure rates under registered-service versus replay/test execution.
\textbf{e,} Mutually exclusive primary causes of the 21 incomplete workflows.
\textbf{f,} Non-recovery after perturbation, unnecessary-call ratio and
cross-case contamination rate. Error bars in \textbf{d} and \textbf{f} show
95\% confidence intervals. Counts and exact percentages are shown for all
endpoints; exact intervals are reported in Supplementary Table~S3.}
\label{fig:e2e}
\end{center}
\clearpage

\begin{center}
\centering
\includegraphics[width=0.86\textwidth,height=0.84\textheight,keepaspectratio]{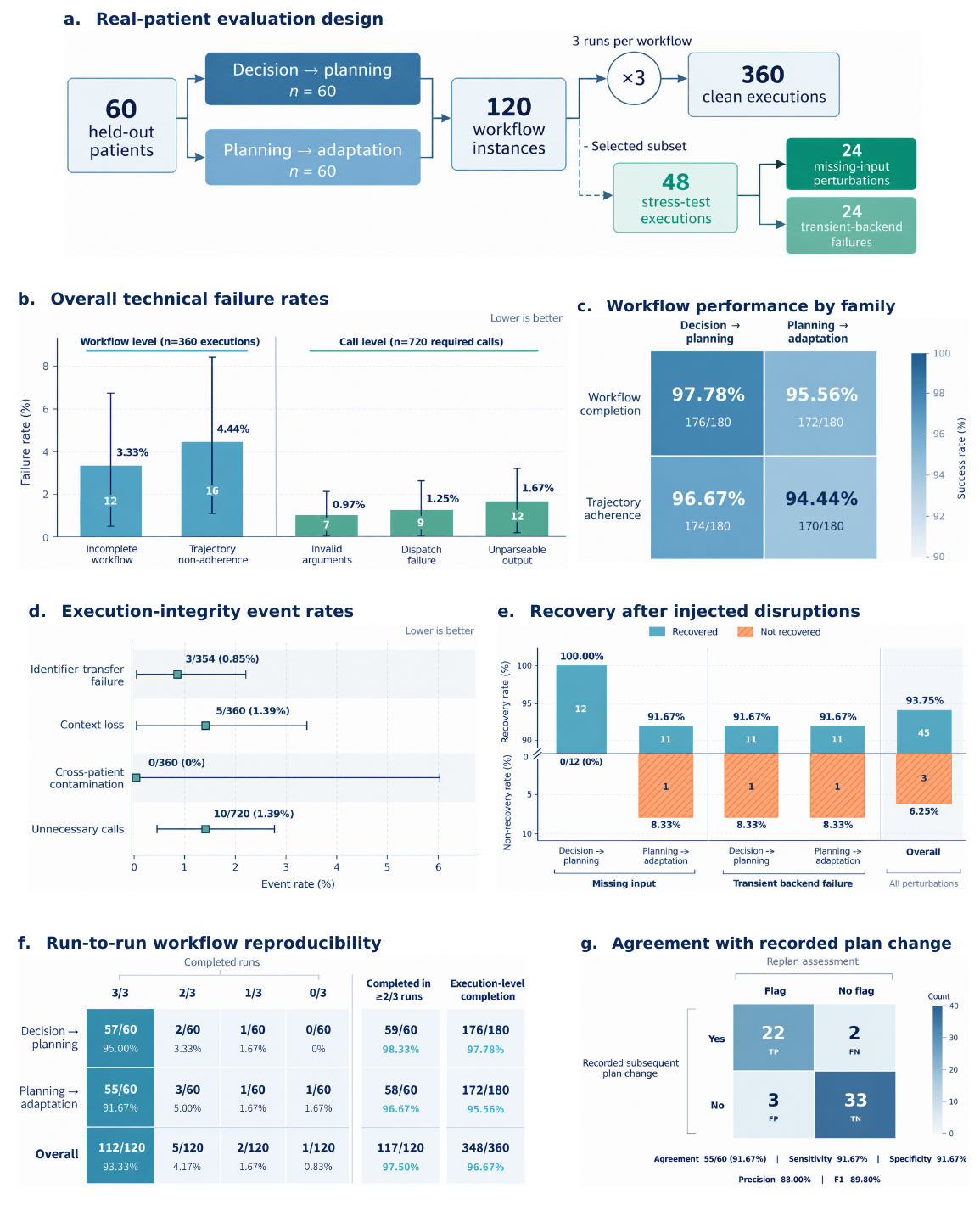}
\captionof{figure}{\textbf{Real-patient workflow-execution evaluation.}
\textbf{a,} Evaluation design for 60 de-identified patients. Each contributed
a decision-to-planning and a planning-to-adaptation workflow (120 instances),
each repeated in three context-reset sessions (360 clean executions); 48
separate stress-test executions assessed recovery. \textbf{b,} Failure rates
for overall workflow- and call-level technical endpoints. \textbf{c,}
Completion and trajectory adherence by workflow family. \textbf{d,} Event
rates for identifier-transfer failure, context loss, cross-patient
contamination and unnecessary calls. \textbf{e,} Recovery and non-recovery
after injected missing-input or transient-backend failures, stratified by
workflow family. \textbf{f,} Workflow completion across zero to three clean
runs, with at-least-two-run and execution-level completion. \textbf{g,}
Confusion matrix and derived performance measures for replan assessment versus
recorded subsequent plan change. Error bars in \textbf{b} and \textbf{d} show
95\% confidence intervals. TP, true positive; FN, false negative; FP, false
positive; TN, true negative. Exact intervals are reported in Supplementary
Table~S3.}
\label{fig:clinical-evaluation}
\end{center}
\clearpage

\begingroup
\setcounter{figure}{0}
\setcounter{table}{0}
\renewcommand{\thefigure}{\arabic{figure}}
\renewcommand{\theHfigure}{EDF\arabic{figure}}
\renewcommand{\figurename}{Extended Data Figure}
\renewcommand{\thetable}{\arabic{table}}
\renewcommand{\theHtable}{EDT\arabic{table}}
\renewcommand{\tablename}{Extended Data Table}
\section*{Extended Data}
\begin{center}
\centering
\includegraphics[width=0.94\textwidth,height=0.78\textheight,keepaspectratio]{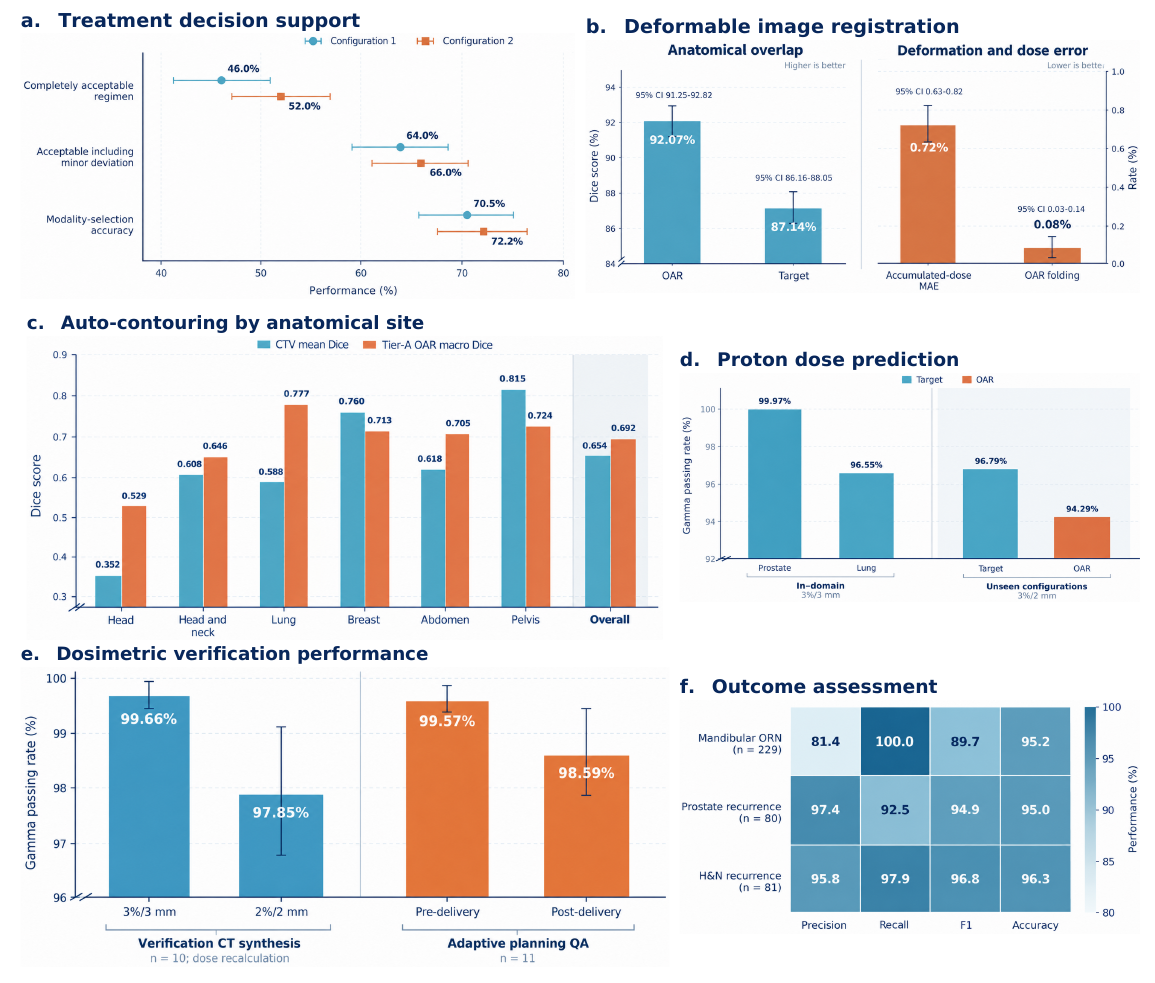}
\captionof{figure}{\textbf{Representative function-specific evidence from the
specialist layer.} Each panel reports function-specific evidence using its
task-appropriate native endpoint. \textbf{a,}
Treatment-regimen acceptability and modality-selection accuracy. \textbf{b,}
OAR and target propagated-contour Dice, accumulated-dose error and OAR folding
for deformable registration. \textbf{c,} Mean CTV Dice and Tier-A OAR macro
Dice across anatomical sites; OAR values are unweighted macro-averages across
the site--organ units. \textbf{d,} In-domain target gamma passing rates at
3\%/3~mm and target and OAR rates for unseen beam configurations or
prescriptions at 3\%/2~mm. \textbf{e,} Dose-recalculation gamma passing rates
for verification-CT synthesis and pre- and post-delivery quality-assurance
rates for adaptive planning. Error bars in \textbf{a}, \textbf{b} and
\textbf{e} denote 95\% confidence intervals. \textbf{f,} Precision, recall,
F1 score and accuracy for physician-adjudicated outcome assessments; H\&N,
head and neck. Complete
function-specific cohorts, reference standards and evidence are provided in
Supplementary Tables~S1 and S2; exact 95\% confidence intervals are
reported in Supplementary Table~S5.}
\label{fig:specialist-evidence}
\end{center}
\clearpage
\begin{landscape}
\captionsetup{font=footnotesize,skip=4pt}
\clearpage
\fontsize{6.6}{7.4}\selectfont
\setlength{\tabcolsep}{1.6pt}
\renewcommand{\arraystretch}{1.04}
\begin{longtable}{>{\raggedright\arraybackslash}p{3.15cm}
                  >{\raggedright\arraybackslash}p{2.55cm}
                  >{\raggedright\arraybackslash}p{3.35cm}
                  >{\raggedright\arraybackslash}p{8.25cm}
                  >{\raggedright\arraybackslash}p{5.10cm}}
\caption{\textbf{Function-to-workflow coverage and documented benchmark exposure.}
Workflow-family abbreviations are D--P, decision-to-planning; P--A,
planning-to-adaptation; T--F, treatment-to-follow-up; and FP, full pathway.
For cross-cutting utilities, ``conditional'' indicates availability when
required by the scenario rather than a mandatory call. SF reports the exact
per-function exposure in the single-function benchmarks: 300 normal-request
executions plus 60 missing-input executions. RP reports required calls in the
360 clean real-patient executions. The cross-stage benchmark contained 3,150
required calls overall, but the available aggregate record does not support
retrospective allocation of this total to individual functions; cross-stage
entries therefore report registered workflow-family roles rather than
per-function call counts.}
\label{tab:function-workflow-coverage}\\
\toprule
\rowcolor{tablehead}
\textbf{Function} & \textbf{Clinical role} & \textbf{Workflow-family role} &
\textbf{Principal cross-function connection} & \textbf{Documented benchmark exposure} \\
\midrule
\endfirsthead
\multicolumn{5}{l}{\tablename~\thetable\ continued}\\
\toprule
\rowcolor{tablehead}
\textbf{Function} & \textbf{Clinical role} & \textbf{Workflow-family role} &
\textbf{Principal cross-function connection} & \textbf{Documented benchmark exposure} \\
\midrule
\endhead
\midrule
\multicolumn{4}{r}{Continued on next page}\\
\endfoot
\bottomrule
\endlastfoot

RT-indication assessment
& Diagnosis, Patient Evaluation \& Decision-Making
& D--P; FP
& Patient context and retrieved guidance $\rightarrow$ treatment decision and planning directive
& SF: 300 normal + 60 missing-input; RP: -- \\

\rowcolor{tablesubtotal}
Modality selection
& Diagnosis, Patient Evaluation \& Decision-Making
& D--P; FP
& Clinical context and candidate-plan information $\rightarrow$ selected modality and planning directive
& SF: 300 + 60; RP: -- \\

\midrule
Planning directive
& Simulation \& Planning
& D--P; FP
& Treatment decision, intent and prescription $\rightarrow$ simulation, contouring and planning inputs
& SF: 300 + 60; RP: 180 required clean calls \\

\rowcolor{tablesubtotal}
MR-to-CT synthesis
& Simulation \& Planning
& D--P; FP
& Planning MRI and body mask $\rightarrow$ synthetic CT for contouring and dose functions
& SF: 300 + 60; RP: -- \\

Deformable registration
& Simulation \& Planning
& P--A; FP
& Reference and verification images $\rightarrow$ propagated anatomy and replan assessment
& SF: 300 + 60; RP: 180 required clean calls \\

\rowcolor{tablesubtotal}
Auto-contouring
& Simulation \& Planning
& D--P; FP
& Planning image and directive $\rightarrow$ targets and OARs for dose prediction or optimization
& SF: 300 + 60; RP: 180 required clean calls \\

Proton dose prediction
& Simulation \& Planning
& D--P; FP
& CT, structures, beams and prescription $\rightarrow$ predicted dose for plan evaluation
& SF: 300 + 60; RP: -- \\

\rowcolor{tablesubtotal}
Plan optimization
& Simulation \& Planning
& D--P; FP
& Directive, contours and planning feedback $\rightarrow$ optimized plan state
& SF: 300 + 60; RP: -- \\

\midrule
Verification CT synthesis
& On-Treatment Adaptation
& P--A; FP
& On-treatment CBCT or kV images $\rightarrow$ CT-like image for registration and adaptive assessment
& SF: 300 + 60; RP: -- \\

\rowcolor{tablesubtotal}
Replan assessment
& On-Treatment Adaptation
& P--A; FP
& Registered anatomy and available dose $\rightarrow$ change metrics and clinician review flag
& SF: 300 + 60; RP: 180 required clean calls \\

\midrule
Completion note
& Follow-up \& Outcomes
& T--F; FP
& Delivered course and adaptation record $\rightarrow$ structured treatment summary for follow-up
& SF: 300 + 60; RP: -- \\

\rowcolor{tablesubtotal}
Recurrence assessment
& Follow-up \& Outcomes
& T--F; FP
& Treatment history, imaging and longitudinal records $\rightarrow$ recurrence assessment
& SF: 300 + 60; RP: -- \\

Mandibular ORN assessment
& Follow-up \& Outcomes
& T--F; FP
& Head-and-neck treatment, dose, imaging and records $\rightarrow$ documented ORN assessment
& SF: 300 + 60; RP: -- \\

\rowcolor{tablesubtotal}
Brain-necrosis assessment
& Follow-up \& Outcomes
& T--F; FP
& Cranial treatment, dose, imaging and records $\rightarrow$ brain-necrosis assessment
& SF: 300 + 60; RP: -- \\

Cardiac-injury assessment
& Follow-up \& Outcomes
& T--F; FP
& Thoracic treatment, cardiac dose and longitudinal records $\rightarrow$ cardiac-injury assessment
& SF: 300 + 60; RP: -- \\

\rowcolor{tablesubtotal}
Radiation esophagitis assessment
& Follow-up \& Outcomes
& T--F; FP
& Thoracic treatment and longitudinal symptoms $\rightarrow$ evidence summary and CTCAE grade
& SF: 300 + 60; RP: -- \\

Radiation pneumonitis assessment
& Follow-up \& Outcomes
& T--F; FP
& Thoracic treatment, imaging and longitudinal symptoms $\rightarrow$ evidence summary and CTCAE grade
& SF: 300 + 60; RP: -- \\

\midrule
\rowcolor{tablesubtotal}
Patient status
& Cross-cutting utility
& Conditional: all families
& Patient, diagnosis and treatment status $\rightarrow$ decision, planning, adaptation or follow-up context
& SF: 300 + 60; RP: -- \\

Implant retrieval
& Cross-cutting utility
& Conditional: all families
& Documented implants and hardware $\rightarrow$ simulation, imaging or planning context
& SF: 300 + 60; RP: -- \\

\rowcolor{tablesubtotal}
Diagnostic imaging
& Cross-cutting utility
& Conditional: all families
& Imaging studies, metadata and reports $\rightarrow$ decision, adaptation or outcome assessment
& SF: 300 + 60; RP: -- \\

In-Basket messages
& Cross-cutting utility
& Conditional: all families
& Patient communications $\rightarrow$ current-status or follow-up context
& SF: 300 + 60; RP: -- \\

\rowcolor{tablesubtotal}
Trial search
& Cross-cutting utility
& Conditional: all families
& Diagnosis and treatment context $\rightarrow$ candidate trials and stored eligibility criteria
& SF: 300 + 60; RP: -- \\

Schedule/conflict checking
& Cross-cutting utility
& Conditional: all families
& Appointments $\rightarrow$ workflow schedule and deterministic conflict report
& SF: 300 + 60; RP: -- \\

\rowcolor{tablesubtotal}
Platform guidance
& Cross-cutting utility
& Conditional: all families
& User objective $\rightarrow$ registered function and limitation guidance
& SF: 300 + 60; RP: -- \\

PubMed search
& Cross-cutting utility
& Conditional: all families
& Clinical question $\rightarrow$ publication and citation metadata
& SF: 300 + 60; RP: -- \\

\rowcolor{tablesubtotal}
PSTAR lookup
& Cross-cutting utility
& Conditional: D--P; P--A; FP
& Material and proton energy $\rightarrow$ stopping-power and range values for planning calculations
& SF: 300 + 60; RP: -- \\

\end{longtable}

\clearpage
\fontsize{8.2}{9.4}\selectfont
\setlength{\tabcolsep}{3.5pt}
\renewcommand{\arraystretch}{1.22}
\begin{longtable}{>{\raggedright\arraybackslash}p{4.00cm}
                  >{\raggedright\arraybackslash}p{4.80cm}
                  >{\raggedright\arraybackslash}p{7.30cm}
                  >{\raggedright\arraybackslash}p{5.30cm}}
\caption{\textbf{Comparative architecture-ablation evaluation.}
All configurations used the same locked benchmark inputs and scoring rules.
The embedding router was assessed only for function selection; the
validation-gate ablation was assessed only through replay or test interfaces.
Values are percentages unless otherwise stated.}
\label{tab:architecture-ablation}\\
\toprule
\rowcolor{tablehead}
\textbf{Comparison} & \textbf{Evaluation set} & \textbf{Primary endpoints} &
\textbf{Results} \\
\midrule
\endfirsthead
\multicolumn{4}{l}{\tablename~\thetable\ continued}\\
\toprule
\rowcolor{tablehead}
\textbf{Comparison} & \textbf{Evaluation set} & \textbf{Primary endpoints} &
\textbf{Results} \\
\midrule
\endhead
\bottomrule
\endlastfoot

Full controller versus embedding-based router
& 2,600 unique single-function requests, each repeated three times
& Intended-function selection overall and by direct, clinical-context,
paraphrased and confusable request type
& Overall: 98.79\% versus 95.49\% (+3.30 pp); direct: +1.28 pp; clinical
context: +2.18 pp; paraphrased: +4.04 pp; confusable: +7.31 pp \\
\midrule

\rowcolor{tablesubtotal}
Full controller versus stateless controller
& 200 cross-stage scenarios, each repeated three times
& Completion, trajectory adherence, information transfer, context retention
and excess-call ratio
& Completion: 96.50\% versus 84.00\%; trajectory: 95.00\% versus 79.50\%;
transfer: 98.31\% versus 88.98\%; context: 99.63\% versus 91.29\%; excess
calls: 1.08\% versus 4.06\% \\
\midrule

Validation gates enabled versus bypassed
& 520 missing-input requests, each repeated three times, and 360
identifier-integrity replay/test executions
& Unsafe missing-input dispatch, mismatch detection, safe rejection and
mismatched dispatch
& Unsafe dispatch: 0\% versus 2.82\%; detection: 98.89\% versus 5.28\%; safe
rejection: 98.33\% versus 4.72\%; mismatched dispatch: 0\% versus 95.28\% \\

\end{longtable}

\clearpage
\fontsize{8.0}{9.2}\selectfont
\setlength{\tabcolsep}{3pt}
\renewcommand{\arraystretch}{1.18}
\begin{longtable}{>{\raggedright\arraybackslash}p{3.60cm}
                  >{\raggedright\arraybackslash}p{6.50cm}
                  >{\raggedright\arraybackslash}p{7.20cm}
                  >{\raggedright\arraybackslash}p{3.50cm}}
\caption{\textbf{Workflow-execution safety controls and fail-safe behavior.}
The controls govern function calling, backend routing and session isolation;
they do not establish the clinical correctness or clinical appropriateness of
specialist-function outputs.}
\label{tab:safety-controls}\\
\toprule
\rowcolor{tablehead}
\textbf{Execution risk} & \textbf{Detection mechanism} &
\textbf{Fail-safe behavior} & \textbf{Control mode} \\
\midrule
\endfirsthead
\multicolumn{4}{l}{\tablename~\thetable\ continued}\\
\toprule
\rowcolor{tablehead}
\textbf{Execution risk} & \textbf{Detection mechanism} &
\textbf{Fail-safe behavior} & \textbf{Control mode} \\
\midrule
\endhead
\bottomrule
\endlastfoot

Missing required input
& Function-specific schema checks identify absent required arguments before
backend routing.
& Backend dispatch is blocked; the agent issues a clarification or withholding
response and can resume the workflow only after the required information is
available.
& Deterministic gate with LLM-mediated clarification \\

\rowcolor{tablesubtotal}
Unregistered function, malformed argument or invalid route
& The function registry validates the function name, required and optional
arguments, and registered backend route.
& The call is rejected before backend invocation rather than routed to an
unregistered or incompatible service.
& Deterministic \\

Patient, course or plan identifier mismatch
& Identifiers in each call are checked against the active session; structured
outputs retain patient-, course- and plan-level provenance.
& The inconsistent call is rejected and is not merged into the active
patient/workflow context.
& Deterministic \\

\rowcolor{tablesubtotal}
Cross-patient context carryover
& Session state is bound to one active patient and treatment course, with no
persistent conversational memory or cross-patient cache; state is cleared
between independent executions.
& Information from a different session is not retained as active patient or
workflow context for subsequent calls.
& Deterministic session isolation \\

Transient backend or asynchronous-job failure
& Backend status, transport errors and asynchronous-job progress are monitored
by the dispatch layer.
& Workflow state is retained for bounded retry or resumption after service
restoration; an unresolved failure is returned explicitly rather than treated
as a successful result.
& Hybrid deterministic handling and LLM-mediated resumption \\

\end{longtable}

\end{landscape}
\endgroup
\clearpage

\end{document}